\documentclass[journal]{IEEEtran}
\newif\ifdraft
\draftfalse

\ifdraft
\newcommand{\todo}[1]{~\textbf{\color{red}{{#1}}}}
\else
\newcommand{\todo}[1]{}
\fi

\usepackage{amsmath,amsfonts}
\usepackage{pifont}
\usepackage{cite}
\usepackage{algorithmic}
\usepackage{algorithm}
\usepackage{array}
\usepackage{siunitx}
\usepackage[caption=false]{subfig}
\usepackage{textcomp}
\usepackage{stfloats}
\usepackage{url}
\usepackage{verbatim}
\usepackage{graphicx}
\usepackage{multirow}
\usepackage{xcolor}
\usepackage{booktabs}
\usepackage{xspace}

\def\BibTeX{{\rm B\kern-.05em{\sc i\kern-.025em b}\kern-.08em
    T\kern-.1667em\lower.7ex\hbox{E}\kern-.125emX}}
\usepackage{balance}

\newcommand{\ours}{\textit{EnergyDiff}\xspace}
\newcommand{\oursabbr}{\textit{E.Diff.}\xspace}
\newcommand{\oursabbrcal}{\textit{E.Diff.C.}}

\newcommand{\kw}[1]{\SI{#1}{\kilo\watt}}

\definecolor{tudelft-warm-purple}{cmyk}{0.58,1,0,0.02}
\newcommand{\changed}[1]{\textcolor{black}{#1}}
\newcommand{\changedii}[1]{\textcolor{black}{#1}}

\DeclareMathOperator{\pe}{PE}
\DeclareMathOperator{\silu}{SiLU}
\DeclareMathOperator{\block}{Block}
\DeclareMathOperator{\attention}{Attention}
\DeclareMathOperator{\softmax}{softmax}
\newcommand{\transpose}{\text{T}}
\DeclareMathOperator{\mha}{MHA}
\DeclareMathOperator{\layernorm}{LayerNorm}
\DeclareMathOperator{\ffn}{FFN}

\newcommand{\tripleast}{$\vcenter{\hbox{\normalfont ***}}$\xspace}

\begin{document}
\title{EnergyDiff: Universal Time-Series Energy Data Generation using Diffusion Models}
\author{Nan Lin,~\IEEEmembership{Student Member,~IEEE}, 
Peter~Palensky,~\IEEEmembership{Senior Member,~IEEE},\\
Pedro~P.~Vergara,~\IEEEmembership{Senior Member,~IEEE}
\thanks{This work used the Dutch national e-infrastructure with the support of the SURF Cooperative using grant no. EINF-6262. Nan Lin is funded by NWO Align4Energy Project NWA.1389.20.251.}%
\thanks{Nan Lin, Peter Palensky, and Pedro P. Vergara are with the Intelligent Electrical Power Grids (IEPG) Group, Delft University of Technology, 2628 CD Delft, The Netherlands (e-mail:\{N.Lin,  P.P.VergaraBarrios, P.Palensky\}@tudelft.nl).}%
\thanks{\changed{The code for models and experiments used in this paper can be found in Github Repositories: https://github.com/sentient-codebot/EnergyDiff-pub and https://github.com/distributionnetworksTUDelft/EnergyDiff-pub}}
}

\markboth{Journal of \LaTeX\ Class Files,~Vol.~18, No.~9, June~2024}%
{Nan Lin: Universal Time-series Energy Data Synthesis Using Denoising Diffusion Probabilistic Models}

\maketitle

\begin{abstract}
High-resolution time series data are crucial for the operation and planning of energy systems such as electrical power systems and heating systems. \changed{Such data often cannot be shared due to privacy concerns, necessitating the use of synthetic data. However, high-resolution time series data is difficult to model due to its inherent high dimensionality and complex temporal dependencies.} \changed{Leveraging the recent development of generative AI, especially diffusion models}, we propose \ours, a universal data generation framework for energy time series data. \ours builds on state-of-the-art denoising diffusion probabilistic models, utilizing a proposed denoising network dedicated to high-resolution time series data and introducing a novel Marginal Calibration technique. Our extensive experimental results demonstrate that \ours achieves significant improvement in capturing the temporal dependencies and marginal distributions compared to baselines, particularly at the 1-minute resolution. \changedii{\ours's universality is validated across diverse energy domains (e.g., electricity demand, heat pump, PV), multiple time resolutions (1 minute, 15 minutes, 30 minutes and 1 hour), and at both customer and transformer levels.}

\end{abstract}


\section{Introduction}
\IEEEPARstart{T}{he} rapid increase in the integration of renewable energy sources into energy systems has resulted in unprecedented volatility in energy generation. Additionally, the electrification of the energy systems has drastically altered energy consumption behaviors. Together, these factors pose significant challenges to energy systems' economical and safe operation and planning. To develop effective operation and planning solutions, energy system operators require accurate energy generation and consumption profiles~\cite{vemalaiah2024SynergisticDayaheadScheduling}, necessitating the collection of large amounts of high-resolution energy time series data. However, collecting such data is challenging due to privacy and cost concerns. Therefore, the need for algorithms that generate realistic energy time series data is crucial. 

\changed{To prepare the energy system for the increasing penetration of low-carbon technologies, detailed energy system data is required. Such data can raise the customer's awareness of their power consumption characteristics, leading to energy cost savings and facilitating retrofitting investments. For the energy system operators, synthetic energy data is crucial for performing future infrastructure safety assessment~\cite{duque2023RiskAwareOperatingRegions} and necessary for reinforcement learning-based operation and planning tasks~\cite{hou2025RLADNHighperformanceDeep}. More broadly, data generation plays an important role in machine learning (ML) tasks involving small and unbalanced datasets. The conventional approach begins with training a generative model on the actual datasets. Subsequently, a large number of synthetic data samples are generated, and these are combined with the real data to train another ML model addressing the specific task.}

Conventional methods, such as Gaussian Mixture Models (GMMs) and t-Copula~\cite{duque2021ConditionalMultivariateElliptical, sun2019LearningVineCopula} model, have been widely used for data generation due to their simplicity and historical effectiveness. In particular, t-Copula has the unique advantage of fitting the marginal distributions precisely, which is suitable for representing high-consumption or high-generation scenarios. However, these models struggle to capture the complex dependencies inherent in high-resolution energy data, leading to sub-optimal performance in representing real-world scenarios. 

Deep generative models offer more advanced solutions by capturing the intricate temporal patterns within the data. Examples are Generative Adversarial Networks (GANs)~\cite{song2022ProfileSRGANGANBased, ma2023AttentionbasedCycleconsistentGenerative}, Variational Auto-encoders (VAEs)~\cite{pan2019DataDrivenEVLoad}, and flow-based models~\cite{xia2024FlowbasedModelConditional}. Despite their strength, data generated by VAEs face challenges in maintaining high-resolution details; GANs suffer from mode collapse and training instability~\cite{goodfellow2014GenerativeAdversarialNets, arjovsky2017PrincipledMethodsTraining}. Despite flow-based models having gained success in recent years~\cite{xia2024FlowbasedModelConditional}, they intrinsically require an invertible neural network structure, limiting its expressing power. More importantly, little effort has been made to generate high-resolution time series data, i.e., 1-minute resolution or higher. For example, a daily electricity consumption profile has 1440 steps at the 1-minute resolution. Such high resolutions pose a great challenge to any generative models. \changed{Take a simple Gaussian model as an example, whose parameters are the mean vector and covariance matrix. Fitting such a model on 1440-step times series data requires finding a large $1440 \times 1440$ covariance matrix, containing more than a million parameters. Consequently, high-resolution models become parameter-heavy, and the fitting process can become unstable. As we will show, the t-Copula model, despite being an efficient and commonly adopted model, faces severe instability issues when applied to high-resolution data because it requires the inversion of a matrix that becomes too large at high resolutions.}

Denoising diffusion probabilistic models (DDPMs) are newly emerged deep generative models, which are easy to train and exhibit state-of-the-art data generation quality and diversity~\cite{ho2020DenoisingDiffusionProbabilistic, dhariwal2021DiffusionModelsBeat}, overcoming the disadvantages of previous deep generative models. These advantages make DDPM a natural candidate for a universal generative energy time series data model. 
Nevertheless, the state-of-the-art DDPM~\cite{peebles2023ScalableDiffusionModels} was designed for image generation and faces several challenges when applied to energy time series generation. The first challenge is how to deal with high-resolution data, such as 1-minute data, as the computation complexity grows rapidly with the time series length due to the Transformer network architecture~\cite{vaswani2017AttentionAllYou}. The second challenge is the inaccurate approximation of the marginal distributions. In the image field, the marginal distribution is the brightness distribution of pixels, which need not be extremely precise. However, in the energy field, the marginal distributions are important for characterizing the peak consumption and generation values. Recently, DDPM has been adopted to model electricity load profiles 
and electric vehicle (EV) charging scenarios~\cite{li2024DiffChargeGeneratingEV}. The model proposed in \cite{li2022EnergyDataGeneration} focuses more on a generative forecast problem, while \cite{li2024DiffChargeGeneratingEV} investigates a shorter time series, i.e., 720 steps, and circumvents the complexity issue by adopting a Long Short-Term Memory (LSTM) network architecture instead of a Transformer. Neither of these works has fully addressed the previous two challenges. 

In this work, we propose \ours, a universal energy time series data generation framework based on DDPM, which is applicable across various energy domains, multiple time resolutions, and at both customer and electrical transformer levels. Furthermore, we propose a simple yet novel Marginal Calibration technique to combine the underlying dependency structure of DDPM and the empirical cumulative distribution functions (CDFs) of training samples, yielding almost exact marginal distributions on any model. Our contributions are summarized as follows.
\begin{itemize}
    \item We propose \ours, a DDPM-based framework that is dedicated to generating energy time-series data. The proposed \ours is 1) scalable across different time resolutions and 2) applicable to generate data at both the \changed{electrical} transformer level and customer (household) level. 
    \item To overcome the limitations of standard DDPM of modeling high-resolution time series data, we propose to use a folding operation in DDPM that quintessentially enables us to generate high-resolution data such as 1 minute with less computation than without the operation. 
    \item We propose a Marginal Calibration technique that calibrates the inaccurate DDPM marginal distributions while preserving the learned complex temporal dependency structure. The proposed technique allows us to use prior knowledge about the marginal distributions in deep generative models. The generated data show significant improvement in terms of Kullback-Leibler divergence, Wasserstein distance, and Kolmogorov–Smirnov statistic.
\end{itemize}

\section{Denoising Diffusion Probabilistic Models}
\begin{figure}[!t]
    \centering
    \includegraphics[width=0.9\linewidth]{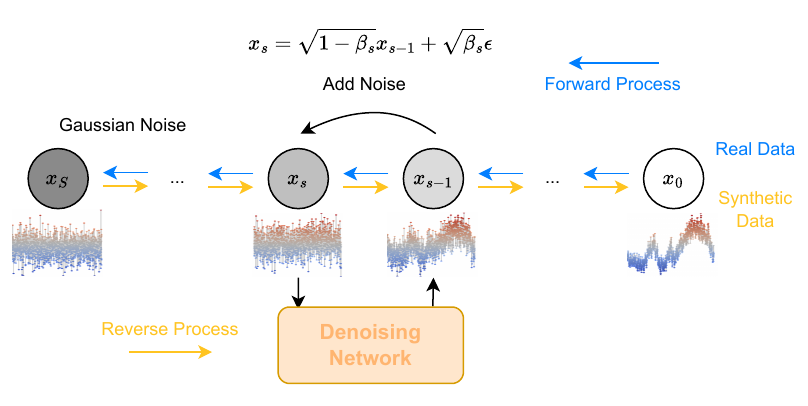}
    \caption{In the forward process, noise is injected into the true data at gradually increasing levels; in the reverse process, the noise is iteratively removed from noisy data. In the proposed framework, we propose a denoising network using customized architecture (see Fig.~\ref{fig:denoising-network}) based on the Transformer network~\cite{vaswani2017AttentionAllYou}. \changedii{The curves under each $x_s$ are examples of the $x_s$. The color indicates their value, red being a higher value and blue a lower value. $x_0$ is a real daily load consumption profile. As $s$ increases, $x_S$ ultimately becomes pure Gaussian noise.}}
    \label{fig:transformer-diffusion}
\end{figure}

In this section, we introduce both the general theoretical framework of DDPM and the practical procedure of training and generation with DDPM. First, we formulate the probabilistic model by constructing two Markov chains. Next, we derive a loss function from the probabilistic model that can be used for efficient training with stochastic gradient descent (SGD). Finally, we show step-by-step how to generate new samples with a trained DDPM. A simple demonstration of DDPM procedure is shown in Fig.~\ref{fig:transformer-diffusion}.

\subsection{Time Series Probabilistic Model}
Any univariate time series data of $T$ steps can be seen as a random vector $\boldsymbol{x} \in \mathbb{R}^T$. We assume it follows an unknown joint distribution $\boldsymbol{x} \sim p(\boldsymbol{x})$. Specifically, in energy time series data generation, $\boldsymbol{x}$ represents a one-day consumption or generation profile. Consequently, the value of $T$ changes with the time resolution. For hourly resolution, $T=24$, while for 1-minute resolution, $T=1440$. 

\subsection{Diffusion Probabilistic Model Formulation}
Overall, the idea of DDPM is to gradually corrupt data and learn how to recover the corrupted data through step-by-step denoising. To be consistent with the DDPM literature, we define $\boldsymbol{x}_0\equiv \boldsymbol{x}$ and the dimensionality $d\equiv T$. The subscript $0$ represents the diffusion step, which will be explained later. With $\boldsymbol{x}_0$ following an unknown distribution $\boldsymbol{x}_0\sim p(\boldsymbol{x}_0)$, DDPM establishes a parametric distribution $p_\theta(\boldsymbol{x}_0)$ to approximate the true data distribution $p(\boldsymbol{x}_0)$. Note that $\theta$ is an abstract collection of all the parameters of the approximate distribution instead of a single parameter. We explain the detailed formulation of $p_\theta(\boldsymbol{x}_0)$ below. 

First, we define a \textit{forward process} that corrupts the data iteratively with Gaussian noise for steps $s=0,1,...,S$, as follows
\begin{align}
\label{eq:forward-process}
    q(\boldsymbol{x}_s|\boldsymbol{x}_{s-1}) &:= \mathcal{N}(\sqrt{1-\beta_s}\boldsymbol{x}_{s-1}, \beta_s\textbf{I}), 
\end{align}
where $\boldsymbol{x}_0$ is our observed data, and $\beta_s \in (0,1)$ is the corruption strength for diffusion step $s$, which usually increases gradually with $s$. $\beta_s$ is a small number so that we do not corrupt the data too fast. The design for $\{\beta_s\}$ is referred to as \textit{noise schedule} in the literature~\cite{ho2020DenoisingDiffusionProbabilistic, nichol2021ImprovedDenoisingDiffusion}. \changed{Notably, thanks to the Gaussian form in~\eqref{eq:forward-process}, we can write the conditional distribution of $\boldsymbol{x}_s$ in closed-form}
\begin{equation}
    q(\boldsymbol{x}_s|\boldsymbol{x}_0) = \mathcal{N}(
        \sqrt{\bar{\alpha}_s} \boldsymbol{x}_0, (1-\Bar{\alpha}_s)\textbf{I}
    ),
    \label{eq:forward-conditional}
\end{equation}
where $\alpha_s:= 1-\beta_s$, $\bar{\alpha}_s := \Pi_{\tau=1}^s{\alpha_\tau}$. \changed{Equation~\eqref{eq:forward-conditional} suggests we can directly jump from $0$ to any diffusion step $s>0$ in the forward process without running through the Markov chain for $s$ steps.}
Here $\bar{\alpha}_s$ can be intuitively seen as the signal strength of $\boldsymbol{x}_s$. Because $\beta_s$ is between $0$ and $1$, $\bar{\alpha}_s$ monotonically decreases. Consequently, when the final step $S$ is large enough, the data will be almost completely corrupted; i.e., $q(\boldsymbol{x}_S|\boldsymbol{x}_{0}) \approx q(\boldsymbol{x}_S) = \mathcal{N}(\textbf{0}, \textbf{I})$. 
In other words, we completely undermine the data with standard Gaussian noise at the last step of the forward process~$S$.

Now, we shift our focus to the \textit{reverse process}, where we start with the fully corrupted data $\boldsymbol{x}_S$ to get the original data $\boldsymbol{x}_0$. If we know the exact denoising distribution $q(\boldsymbol{x}_{s-1}|\boldsymbol{x}_s)$, for any $s$, we can sample $\boldsymbol{x}_S\sim \mathcal{N}(\boldsymbol{0},\textbf{I})$ and go through the forward process in reverse direction to obtain $\boldsymbol{x}_0$. However, $q(\boldsymbol{x}_{s-1}|\boldsymbol{x}_s)$ is not tractable. Naturally, we can approximate it with the following distribution parameterized by $\theta$
\begin{align}
    p_{\theta}(\boldsymbol{x}_{s-1}|\boldsymbol{x}_s) &:= \mathcal{N}(\boldsymbol{\mu}_{\theta}(\boldsymbol{x}_s,s), \boldsymbol{\Sigma}_\theta(\boldsymbol{x}_s,s)).
\end{align}
The two functions $\boldsymbol{\mu}_\theta$ and $\boldsymbol{\Sigma}_\theta$ tell us how we can denoise $\boldsymbol{x}_s$ to get the less noisy $\boldsymbol{x}_{s-1}$.

\changed{We find $\theta$ by minimizing the divergence between the following two distributions}
\begin{align}
    q(\boldsymbol{x}_{0:S}) &= p(\boldsymbol{x}_0)\Pi_{s=1}^{T}q(\boldsymbol{x}_s|\boldsymbol{x}_{s-1})\\
    p_\theta(\boldsymbol{x}_{0:S}) &= q(\boldsymbol{x}_S)\Pi_{s=1}^{S}p_\theta(\boldsymbol{x}_{s-1}|\boldsymbol{x}_s),
\end{align}
\changed{which are the exact and approximate distribution over $\boldsymbol{x}_{0:S}$.}

\subsection{Training DDPM}
DDPM consists of the forward and reverse processes. The forward process is simply corrupting data and is only a means to an end. Generating data requires only the reverse process, which reduces to evaluating two parametric functions, $\boldsymbol{\mu}_\theta$ and $\boldsymbol{\Sigma}_\theta$. During the training process, we find the parametric functions $\boldsymbol{\mu}_\theta$ and $\boldsymbol{\Sigma}_\theta$ by minimizing a loss function evaluated on a set of training samples $\{\boldsymbol{x}_0^{(i)}\}_{i=1}^N$. 

\subsubsection{Loss Function}
To minimize the discrepancy between the true distribution $p(\boldsymbol{x}_0)$ and the approximate distribution $p_\theta(\boldsymbol{x}_0)$, we can minimize the negative evidence lower bound (ELBO)
\begin{align}
    \mathcal{L}_{\theta} &:= \mathbb{E}_{\boldsymbol{x}_{1:S}\sim q, \boldsymbol{x}_0 \sim p}{\left[-\log{\frac{p_\theta(\boldsymbol{x}_{0:S})}{q(\boldsymbol{x}_{1:S}|\boldsymbol{x}_0)}}\right]} \label{eq:elbo}\\
    &= \mathbb{E}_{\boldsymbol{x}_0\sim p}[-\log{p_\theta(\boldsymbol{x}_0)}
        + D_\textit{KL}(q(\boldsymbol{x}_{1:S}|\boldsymbol{x}_0)||p_\theta(\boldsymbol{x}_{1:S}|\boldsymbol{x}_0))
    ] \nonumber \\
    &\geq \mathbb{E}_{\boldsymbol{x}_0\sim p}{[-\log{p_\theta(\boldsymbol{x}_0)}]}. \nonumber
\end{align}
By optimizing $\mathcal{L}_\theta$, we jointly maximize the log likelihood of data on our model $\log{p_\theta(\boldsymbol{x}_0)}$ and minimizes the approximation error between $p_\theta$ and the true distribution $p(\boldsymbol{x}_0)$. 

Essentially, the parametric functions we need to learn are $\boldsymbol{\mu}_\theta:\mathbb{R}^d\times\mathbb{Z}_{0+} \rightarrow \mathbb{R}^d$ and $\boldsymbol{\Sigma}_\theta:\mathbb{R}^d\times\mathbb{Z}_{0+} \rightarrow \mathbb{S}^d_+$, where $\mathbb{S}^d_+$ is the set of all $d\times d$ positive semi-definite matrices. Naturally, we can use neural networks to parameterize these two functions. Since these two functions essentially serve the purpose of partially removing the noise in $\boldsymbol{x}_s$ to recover $\boldsymbol{x}_{s-1}$, we will refer to them as the \textit{denoising networks}.
The powerful capacity of neural networks can therefore enable us to learn complex joint distributions. As~\cite{ho2020DenoisingDiffusionProbabilistic} suggests, the variance function $\boldsymbol{\Sigma}_\theta$ can be fixed to $\boldsymbol{\Sigma}_\theta=\beta_s\textbf{I}$ with little to no performance drop. Furthermore, for efficient training with SGD, we can derive a loss function from~\eqref{eq:elbo} as
\begin{align}
    &\hat{\mathcal{L}}^{\text{simple}}_\theta = \frac{1}{B}\sum_{i=1}^{B}{
        \frac{1}{2\Tilde{\beta}_{s_i}} ||\tilde{\boldsymbol{\mu}}_{s_i}(\boldsymbol{x}^{(i)}_{s_i},\boldsymbol{x}^{(i)}_0) - \boldsymbol{\mu}_\theta(\boldsymbol{x}^{(i)}_{s_i},s_i)||_2^2
    }, \\
    &\tilde{\boldsymbol{\mu}}_s(\boldsymbol{x}_s,\boldsymbol{x}_0) =          \frac{\sqrt{\bar{\alpha}_{s-1}}\beta_s}{1-\Bar{\alpha}_s} \boldsymbol{x}_0 + \frac{\sqrt{\alpha_s}(1-\bar{\alpha}_s)}{1-\bar{\alpha}_s}\boldsymbol{x}_s
\end{align}
where $\{\boldsymbol{x}_0^{(i)}\}_{i=1}^B$ are a batch of $B$ samples drawn from the complete dataset $\{\boldsymbol{x}_0^{(i)}\}_{i=1}^N$, and each $s_i$ is uniformly and independently drawn from $\{1,...,S\}$. Taking the gradient  $\nabla_\theta\hat{\mathcal{L}}^{\text{simple}}_\theta$ enables us to perform SGD.

\subsubsection{Training Procedures}
Despite the complex construction of DDPM, the training procedure is simple. We summarize it in Algorithm~\ref{alg:train-ddpm}.\todo{this subsubsection feels a bit short, although it is accompanied by an algorithm block.}
\begin{algorithm}[t]
\caption{Training DDPM}
\label{alg:train-ddpm}
\begin{algorithmic}
    \REQUIRE Dataset $\{\boldsymbol{x}_0^{(i)}\}_{i=1}^N$, initialized $\boldsymbol{\mu}_\theta$, learning rate $\eta$, noise schedule $\{\beta_s, \alpha_s, \bar{\alpha}_s\}$
    \WHILE{not converged}
        \STATE Draw samples $\{\boldsymbol{x}^{(i)}_0\}_{i=1}^B$ from the dataset
        \STATE \changed{Randomly choose a} diffusion step $\{s_{i} | s_i \sim \mathcal{U}\{1,S\}\}$ 
        \STATE \textit{Forward} $0\rightarrow s_i$: $\boldsymbol{x}^{(i)}_{s_i} = \sqrt{\Bar{\alpha}_{s_i}}\boldsymbol{x}^{(i)}_0 + \sqrt{1-\Bar{\alpha}_{s_i}}\boldsymbol{\epsilon}_{s_i}$
        \STATE \textit{Reverse} ${s_i}\rightarrow {s_i}-1$: $\boldsymbol{\mu}_\theta(\boldsymbol{x}^{(i)}_{s_i},{s_i})$
        \STATE $\theta \leftarrow \theta + \eta \nabla_\theta{\hat{\mathcal{L}}^{\text{simple}}_\theta(\boldsymbol{x}^{(i)}_{s_i}, \boldsymbol{x}^{(i)}_0, \boldsymbol{\mu}_\theta(\boldsymbol{x}^{(i)}_{s_i},{s_i}))}$
    \ENDWHILE
\end{algorithmic}
\end{algorithm}

\begin{algorithm}[t]
\caption{Sampling new data from DDPM}
\label{alg:sample-ddpm}
\begin{algorithmic}
    \REQUIRE trained $\boldsymbol{\mu}_\theta$, noise schedule $\{\beta_s, \alpha_s, \bar{\alpha}_s\}$
    \STATE Initialize $s = S$
    \STATE Sample $\boldsymbol{x}_S \sim \mathcal{N}(\boldsymbol{0},\textbf{I})$
    \WHILE{$s > 0$}
        \STATE \textit{Reverse}: Calculate $\hat{\boldsymbol{\mu}}_{s-1} = \boldsymbol{\mu}_\theta(\boldsymbol{x}_s,s)$
        \STATE Sample $x_{s-1} \sim \mathcal{N}(\hat{\boldsymbol{\mu}}_{s-1}, \tilde{\beta}_s\textbf{I})$
        \STATE $s \leftarrow s-1$
    \ENDWHILE
    \RETURN $\boldsymbol{x}_0$
\end{algorithmic}
\end{algorithm}

\subsection{Generation Procedure}
Once the training is done and parameters $\theta$ are obtained, we can generate new data through the \textit{reverse process}. As summarized in Algorithm~\ref{alg:sample-ddpm}, we start with sampling noise $\boldsymbol{x}_S$ from a standard normal distribution. Following that, we recursively denoise from $\boldsymbol{x}_s$ to $\boldsymbol{x}_{s-1}$ for $S$ times with the help of denoising distribution $p_{\theta}(\boldsymbol{x}_{s-1}|\boldsymbol{x}_s)$. Finally, we achieve clean sample $\boldsymbol{x}_0$ that approximately follows the true data distribution $p(\boldsymbol{x}_0)$. 

\section{EnergyDiff Architecture}
DDPM is a powerful probabilistic model that can approximate complex distributions. However, there is a notable lack of effort in developing a robust and universally applicable DDPM capable of generating high-resolution energy time series data of various energy domains. There are several challenges in modeling energy time series data by commonly adopted GMM and the standard DDPM. First, the temporal dependencies of energy data vary significantly across different domains and are often complex. Second, the computation and memory complexity can grow dramatically as the time resolution increases. For example, a daily electricity consumption profile with a 1-minute resolution yields a 1440-dimensional vector. This means even a simple Gaussian model with a full covariance matrix would have over a million parameters. Third, neural network-based methods can usually learn complex dependency structures well, but the learned marginal distributions are far less accurate than the empirical cumulative distribution function (ECDF), which is easy to estimate. 

We address all of these challenges in our proposed \mbox{\ours} framework, which is dedicated to energy time series data generation. %
The forward process follows the exact same paradigm as the original DDPM, while the reverse process consists of our tailored denoising network. We also propose an additional Marginal Calibration step upon the completion of the reverse process, which compensates for the inaccuracy of the DDPM on marginal distributions. The complete framework is demonstrated in Fig.~\ref{fig:transformer-diffusion}.

\subsection{Tailored Reverse Process}
\begin{figure}[t]
    \centering
    \includegraphics[width=\linewidth]{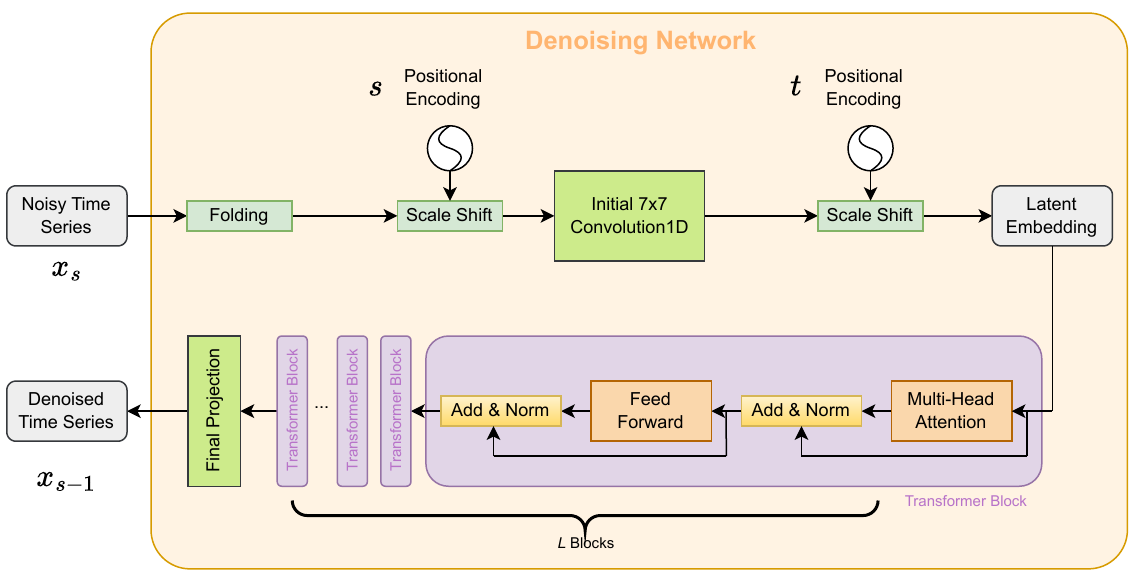}
    \caption{Denoising Network Architecture. The network predicts denoised $\boldsymbol{x}_{s-1}$ based on noisy $\boldsymbol{x}_{s}$, letting us advance one step in the \textit{reverse process}. After the Folding, Initial Convolution, and Positional Encoding blocks, the Transformer blocks serve a central role in denoising data by learning its temporal dependency.} 
    \label{fig:denoising-network}%
    \vspace{-1mm}
\end{figure}

Learning the temporal dependency structure is central in generating the energy time series data. To achieve this, we propose the neural network architecture shown in Fig.~\ref{fig:denoising-network}. The proposed architecture exploits Transformer\footnote{We only refer to the neural network model proposed~\cite{vaswani2017AttentionAllYou} as Transformer in this section.} networks' capacity for processing sequential data (such as time series data)~\cite{vaswani2017AttentionAllYou}. As we will show in Section~\ref{sec:results}, such a design allows \ours to learn complex temporal patterns across different energy domains.
Besides the Transformer blocks, the proposed architecture comprises a Folding block, a two-level Positional Encoding block, an Initial Convolution block, and a Final Projection block.

\subsubsection{Folding} 
\begin{figure}[t]
    \centering
    \includegraphics[width=0.75\linewidth]{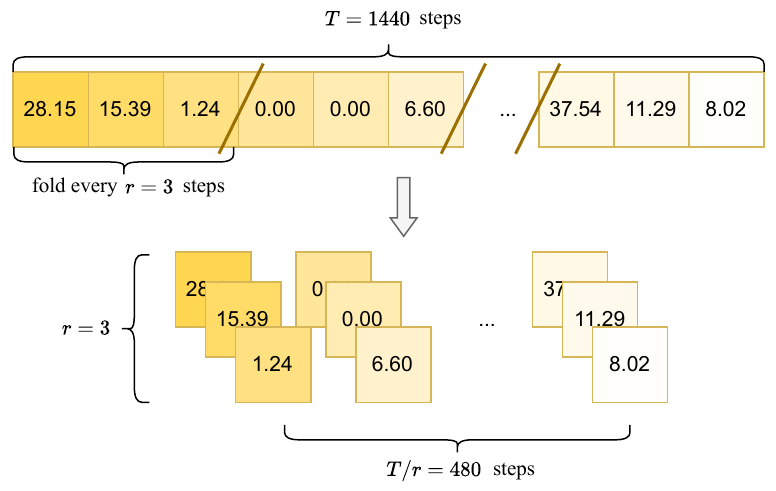}
    \caption{A long univariate time series of 1440 steps is folded into a shorter $3$-variable (or $3$-channel) multivariate time series of 480 steps, with a folding factor $r=3$. }
    \label{fig:folding}%
    \vspace{-1mm}
\end{figure}

Transformer is a powerful model that captures complex temporal patterns. However, its memory and computation complexity is quadratic to the time series length~\cite{vaswani2017AttentionAllYou}. This means whenever time resolution is doubled, we will have four times the computation and memory cost. Sequence length also heavily influences time complexity because of massive memory read/write operations. Therefore, to deal with high-resolution time series, e.g., 1 minute, we propose to use a folding operation as the first step. For a multivariate time series data $\boldsymbol{x}_0 \in \mathbb{R}^{d \times T}$, we fold every consecutive $r$ steps into the channel dimension. This can be represented as
\begin{equation}
    \Tilde{\boldsymbol{x}}_0 \in \mathbb{R}^{dr\times \frac{T}{r}} \leftarrow \boldsymbol{x}_0 \in \mathbb{R}^{d \times T}.
\end{equation}
For long sequences, the computation of Transformers is intensive mostly because of the Attention operation. Specifically, the memory complexity of Attention for sequence length $L$ is nearly $\mathcal{O}(L^2)$. Such a folding operation would reduce the complexity of Transformer operations by $r^2$ times. Fig.~\ref{fig:folding} shows an example of this operation. 

\todo{Critically, the proposed folding operation compromises the inherent data structure to achieve lower complexity. Consequently, the factor $r$ ought to be minimized as much as the complexity permits. However, in practice, the negative impact of the folding operation is mitigated when we use a deeper network. \textbf{optional experiment with different $r$.}}

\subsubsection{Positional Encoding} Before passing data to the Transformer blocks, we need to embed positional information into the data. There are two types of \textit{positions} we need to inform. Generally, for a multivariate sequence at the $s$-th step in the diffusion process, $\boldsymbol{x}_s \in \mathbb{R}^{d \times T}$. The first \textit{position} is $s \in \{0, 1, ..., S\}$, which is implicitly associated with the noise level; the second \textit{position} is $t \in \{0, 1, ..., T-1\}$, the position in the time series. We embed each of them separately through the same mechanism called postional encoding~\cite{vaswani2017AttentionAllYou} and a learned scale and shift. 
\begin{align}
    \pe_{2i}(pos, d) &= \sin\left(\frac{pos}{10000^{2i/d}}\right) \\
    \pe_{2i+1}(pos, d) &= \cos\left(\frac{pos}{10000^{2i/d}}\right),
\end{align}
where $\pe(\cdot, d): \mathbb{N} \rightarrow [-1,1]^{d}$ is a vector function that maps a position $pos$ to a $d$-dimensional vector for any given $d$. Each one element of the $d$-dimensional output has different sensitivities to the change of $pos$. Next we use two fully connected layers to scale and shift $\boldsymbol{x}_s$. For \textit{diffusion step encoding}, we have
\begin{align}
    \boldsymbol{\sigma}(s) &= W^{\text{scale}}_2 \silu(W_1 \pe(s, d) + \boldsymbol{b}_1) + \boldsymbol{b}^{\text{scale}}_2 \\
    \boldsymbol{\delta}(s) &= W^{\text{shift}}_2 \silu(W_1 \pe(s, d) + \boldsymbol{b}_1) + \boldsymbol{b}^{\text{shift}}_2,
\end{align}
where $W$ (matrix) and $\boldsymbol{b}$ (vector) are weights and biases that will be learnt during training, and $\silu$ is the Sigmoid Linear Unit activation function proposed by~\cite{hendrycks2023GaussianErrorLinear}. With $\boldsymbol{\sigma}(s)\in \mathbb{R}^d$ and $\boldsymbol{\delta}(s)\in \mathbb{R}^d$, we scale and shift $\boldsymbol{x}_{s,t}$ by
\begin{equation}
    \boldsymbol{x}_{s,t}  \leftarrow \left(1+\boldsymbol{\sigma}(s)\right) \odot \boldsymbol{x}_{s,t} + \boldsymbol{\delta}(s),
\end{equation}
where $\odot$ is the Hadamard (element-wise) product. The scale and shift are only determined by $s$ and stay the same for $\forall t$ given the same $s$. We use $1+\boldsymbol{\sigma}(s)$ instead of $\boldsymbol{\sigma}(s)$ and initialize $\boldsymbol{\sigma}$ and $\boldsymbol{\delta}$ with zero, as this showed capabilities of stabilizing the training. 

Additionally, another Positional Encoding is added for \textit{time step} $t$, as shown in Fig.~\ref{fig:denoising-network}. This Positional Encoding is placed after Initial Convolution and before the Transformer blocks. After the Initial Convolution, we have sequence $\boldsymbol{x}^\prime_s \in \mathbb{R}^{d^\prime \times T}$. We use the same sinusoidal positional encoding with a different dimensionality $\pe(t, d^\prime)$ and another two fully connected layers applying scale and shift on $\boldsymbol{x}_{s,t}^\prime$. 
\begin{align}
    \boldsymbol{\sigma}^\prime(t) &= W^{\prime \,\text{scale}}_2 \silu(W^\prime_1 \pe(t, d^\prime) + \boldsymbol{b}^\prime_1) + \boldsymbol{b}^{\prime \,\text{scale}}_2 \\
    \boldsymbol{\delta}^\prime(t) &= W^{\prime \,\text{shift}}_2 \silu(W^\prime_1 \pe(t, d^\prime) + \boldsymbol{b}^\prime_1) + \boldsymbol{b}^{\prime\,\text{shift}}_2,
\end{align}
where the $W$ matrices and $\boldsymbol{b}$ vectors are also learnable parameters. Different from above, we scale and shift $\boldsymbol{x}^\prime_{s,t}$ by 
\begin{equation}
    \boldsymbol{x}^\prime_{s,t} \leftarrow (1+\sigma^\prime(t)) \odot \boldsymbol{x}^\prime_{s,t} + \delta^\prime(t).
\end{equation}
The scale and shift is only determined by $t$ and stay the same for $\forall s$. 

\subsubsection{Initial Convolution} After folding and positional encoding for $s$, we employ a large kernel convolution, which is proved useful as an initial feature extractor~\cite{ding2022ScalingYourKernels}. We use the same convolution for $\forall s$. 
\begin{equation}
    \boldsymbol{x}^\prime_{s,t} = \sum_{\tau = -k}^{k}{
        W^{\text{init}}_{\tau}\boldsymbol{x}_{s,t-\tau}
    } + \boldsymbol{b}^{\text{init}},
\end{equation}
where $2k+1$ is the convolution kernel size; $W^{\text{init}}_{\tau} \in \mathbb{R}^{d^\prime \times d}$ and $\boldsymbol{b}^{\text{init}}\in\mathbb{R}^{d^\prime}$ are the learnable weight matrices and bias. Since $t_{\text{min}}=0$, we use a circular padding for $\boldsymbol{x}_{s,t-1},...,\boldsymbol{x}_{s,t-k}$.

\subsubsection{Transformer Blocks} The transformer blocks are our main tool for learning temporal dependency. Since our task is denoising, we do not need the \textit{encoder-decoder} structure in~\cite{vaswani2017AttentionAllYou}. Instead, we only adopt the decoders. This has been proven to work effectively for image generation~\cite{peebles2023ScalableDiffusionModels}. There are two main sub-blocks in a Transformer block, namely multi-head attention (MHA) and feed forward network (FFN). We use a total of $L$ Transformer blocks. For block $l$, it takes the output of previous block $\boldsymbol{x}_{s,t}^{(l-1)}$ as input and passes its output $\boldsymbol{x}_{s,t}^{(l)}$ to the next block. The initial input is $\boldsymbol{x}_{s,t}^{(0)} := \boldsymbol{x}^\prime_{s,t}$, with the final output being $\boldsymbol{x}_{s,t}^{(L)}$. We fix the dimensionality of the input and output such that
\begin{equation}
    \block_l: \boldsymbol{x}_{s,t}^{(l-1)} \in \mathbb{R}^{d^\prime\times T} \rightarrow \boldsymbol{x}_{s,t}^{(l)} \in \mathbb{R}^{d^\prime\times T}, \;\forall l\{1,...,L\}.
\end{equation}

MHA is the core operation of Transformer. It extracts the temporal features by comparing the sequence at each time step with any other time step. An $\mha$ has $H$ heads; each head operates separately and is aggregated later. Using multiple heads allows the attention mechanism to focus on different attributes of the data. For head $h$ in the $l$-th Transformer block, we first calculate
\begin{align}
    \boldsymbol{q}_{l,h} &= W^{\text{q}}_{l,h} \boldsymbol{x}_{s,t}^{(l-1)} \\
    \boldsymbol{k}_{l,h} &= W^{\text{w}}_{l,h} \boldsymbol{x}_{s,t}^{(l-1)} \\
    \boldsymbol{v}_{l,h} &= W^{\text{v}}_{l,h} \boldsymbol{x}_{s,t}^{(l-1)},
\end{align}
where $W^{\text{q}}_{l,h},W^{\text{w}}_{l,h},W^{\text{v}}_{l,h} \in \mathbb{R}^{\frac{d^\prime}{H}\times d^\prime}$ are learnable matrices. $\boldsymbol{q}_{l,h}, \boldsymbol{k}_{l,h}, \boldsymbol{v}_{l,h}$ are called queries, keys, and values respectively. For each of the $T$ query, the idea is to calculate the similarity between the query and each of the $T$ keys with dot product. After normalizing the dot product, we have a series of similarity weights that sum up to one. We then use these weights to perform a weighted average of the associated values. 
\begin{align}
    &\attention_h(\boldsymbol{q}_{l,h},\boldsymbol{k}_{l,h},\boldsymbol{v}_{l,h}) = \boldsymbol{v}_{l,h}\softmax(\frac{\boldsymbol{k}_{l,h}^\transpose \boldsymbol{q}_{l,h}}{\sqrt{{d^\prime}/{H}}}) \\
    &\softmax_{m,n}(\zeta) := \frac{e^{\zeta_{m,n}}}{\sum_{i=1}^{M}{e^{\zeta_{i,n}}}}, \; \zeta \in \mathbb{R}^{M\times N},
\end{align}
where the $\softmax$ function normalizes a matrix by the columns. The sum of any column of the $\softmax$ output is always one. Each attention head operates independently and their outputs are concatenated to get the final output.
\begin{multline}
    \mha(\boldsymbol{x}_{s}^{(l-1)}) \\
        = \text{Concat}(\attention_1, \attention_2, ..., \attention_H),
\end{multline}
where the output has the same shape as the input $\mathbb{R}^{d^\prime\times T}$. 

Following the MHA, we add a $\layernorm$ and a skip connection.
\begin{align}
    &\tilde{\boldsymbol{x}}_{s,t}^{(l-1)} = \layernorm(\boldsymbol{x}_{s,t}^{(l-1)}) + [\mha(\boldsymbol{x}_{s}^{(l-1)})]_t \\
    &\layernorm(\boldsymbol{x}_{s,t}^{(l-1)}) = \frac{\boldsymbol{x}_{s,t}^{(l-1)}-\mathbb{E}[\boldsymbol{x}_{s,t}^{(l-1)}]}{\sqrt{\text{Var}[\boldsymbol{x}_{s,t}^{(l-1)}] + \epsilon}} \odot \boldsymbol{\gamma} + \boldsymbol{\beta},
\end{align}
where $\layernorm$ normalizes $\boldsymbol{x}_{s,t}^{(l-1)} \in \mathbb{R}^{d^\prime}$ over the $d^\prime$ elements. $\epsilon$ is a small number for numerical stability. $\boldsymbol{\gamma}$ and $\boldsymbol{\beta}$ are learnable $\mathbb{R}^{d^\prime}$ vectors. 

Next, we pass $\tilde{\boldsymbol{x}}_{s,t}^{(l-1)}$ to a two-layer feed forward (fully connected) network.
\begin{equation}
    \ffn(\tilde{\boldsymbol{x}}_{s,t}^{(l-1)}) = W^{\text{FF}}_{l,2} \silu(W^{\text{FF}}_{l,1}\tilde{\boldsymbol{x}}_{s,t}^{(l-1)} + \boldsymbol{b}^{\text{FF}}_{l,1}) + \boldsymbol{b}^{\text{FF}}_{l,2},
\end{equation}
where the $W$ matrices and $b$ vectors are learnable weights and biases. Consequently, we get out final output of this layer with another $\layernorm$ and a skip connection.
\begin{equation}
    \boldsymbol{x}_{s,t}^{(l)} = \layernorm(\tilde{\boldsymbol{x}}_{s,t}^{(l-1)}) + \ffn(\tilde{\boldsymbol{x}}_{s,t}^{(l-1)})
\end{equation}

\subsubsection{Final Projection} After $L$ blocks of Transformer, we perform an affine projection on concatenated $x_{s,t}^{(L)}$ and $x_{s,t}^{(0)}$. This serves as a partial skip connection that helps with building deep networks. 
\begin{equation}
    \hat{\boldsymbol{\mu}}_{s-1,t} = W^{\text{o}} \text{Concat}(\boldsymbol{x}_{s,t}^{(0)}, \boldsymbol{x}_{s,t}^{(L)}) + \boldsymbol{b}^{\text{o}},
\end{equation}
where $W^{\text{o}}$ and $b^{\text{o}}$ are learnt parameters. $\hat{\boldsymbol{\mu}}_{s-1}$ is the estimated mean of denoising distribution $p_\theta(\boldsymbol{x}_{s-1}|\boldsymbol{x}_s)=\mathcal{N}(\hat{\boldsymbol{\mu}}_{s-1}, \tilde{\beta}_s\textbf{I})$, and $\theta$ is the collection of all of the learnable parameters above. 

\subsection{Optimal Marginal Calibration}
All joint distributions comprise two elements: the dependency structure and the marginal distributions. 
Estimating the dependency structure is generally challenging, whereas the marginal distributions can be straightforwardly and precisely estimated by methods such as ECDF or simple parametric 1D distributions when prior knowledge is available. Neural network-based models approximate these two elements simultaneously by minimizing the ELBO. In practice, the resulting marginal distributions can have significant discrepancies with the true marginal distributions\todo{cite previous works comparing the marginals}. To address these inaccuracies, we propose a marginal calibration process utilizing optimal transport (OT) mapping. This calibration applies the minimal alterations, maintaining the original temporal dependency structure while aligning the variables with the accurate marginal distributions. \changed{The proposed technique is based on simple numerical methods and can be seen as a post-processing step after generating synthetic data from the previous diffusion model.}

\subsubsection{Re-estimate Marginal Distribution}
First, we acquire a new estimate of the marginal distributions. In the most general case, we exploit the marginal ECDF, which is an unbiased and easily accessible estimate of the true marginal CDF. \changed{The Glivenko–Cantelli theorem states that the ECDF uniformly converges to the true CDF, making the ECDF a good estimate for the unknown true CDF~\cite{glivenko1933sulla, cantelli1933sulla}.} Practically, there is often prior knowledge about the marginal distribution of energy time series data. For example, the marginals of residential electrical energy consumption often follow a log-normal or gamma distribution~\cite{carpaneto2008ProbabilisticCharacterisationAggregated}. In such cases, we can perform more accurate and statistically sound estimations by maximum likelihood or maximum a posteriori (MAP). \changed{This paper takes the general approach and uses ECDF to estimate the marginal distributions.} Formally, to obtain ECDF of the real (training) data, we have
\begin{equation}
    F^*_{t}(\nu) = \frac{1}{N}\sum_{i=1}^{N} \mathbf{1}_{x\leq \nu}(x^{(i)}_{0,t})
\end{equation}
\begin{equation}
    \mathbf{1}_{x\leq \nu}(x^{(i)}_{0,t}) = 
        \begin{cases}
            1 & \text{if $x^{(i)}_{0,t} \leq \nu$} \\
            0 & \text{otherwise}  
        \end{cases}
\end{equation}
where $x^{(i)}_{0,t}$ is the $t$-th time step of the $i$-th sample taken from the training set $\{\boldsymbol{x}^{(i)}_{0}\}_{i=1}^{N}$. $\mathbf{1}_{x\leq \nu}(x)$ is an indicator function that outputs $1$ if the input is less than or equal to $\nu$ and $0$ otherwise. 

Meanwhile, we also estimate the (\changed{potentially inaccurate}) marginal distributions of DDPM. After training our model, we first generate $M$ synthetic samples $\{\hat{x}^{(i)}_{0}\}_{i=1}^{M}$. The ECDF of DDPM is therefore given by
\begin{equation}
    F^\prime_{t}(\nu) = \frac{1}{M}\sum_{i=1}^{M} \mathbf{1}_{\hat{x}^{(i)}_{0,t}\leq \nu},
\end{equation}
where $\hat{x}^{(i)}_{0,t}$ is generated synthetic sample. \changed{Notably, we do the above estimation independently for each time step $t$.}

\subsubsection{Optimal Transport Calibration}
Next, we seek a way to replace the inaccurate marginal distributions $F^\prime_t$ with the more accurate ones $F^*_t$. We find this mapping $g_t$ by solving the following optimization problem
\begin{align}
    \label{eq:ot}
    &\min_{g_t} \mathbb{E}_{\hat{x}_{0,t}\sim F_t^\prime}[||\hat{x}_{0,t} - g_t(\hat{x}_{0,t})||_2^2] \\
    &s.t. \;\;\forall \hat{x}_{0,t}\sim F_t^\prime, \;g_t(\hat{x}_{0,t})\sim F_t. \nonumber
\end{align}
The constraints indicate that the mapping $g_t$ must transform a random variable from the distribution $F^\prime_t$ into a random variable that conforms to the new distribution $F_t$. The objective implies that we seek a mapping $g_t$ close to the identity mapping, as it minimizes $\mathbb{E}_{\hat{x}_{0,t}\sim F_t^\prime}[||\hat{x}_{0,t} - g_t(\hat{x}_{0,t})||_2^2]$. 

This problem is also known as the OT problem. The exact solution is called OT mapping. In general, the OT problem is a complex functional problem. \changed{However, in the one-dimensional case, the solution is explicitly given by~\cite{rachev1998MassTransportationProblems}.} For~\eqref{eq:ot}, the exact OT mapping $g_t^*$ is given by
\begin{equation}
    g_t^*(\hat{x}_{0,t}) = F_t^{*-1}(F_t^{\prime}(\hat{x}_{0,t})) \;\;\forall t.
\end{equation}
\changed{Repeating the above calculations $T$ times for each $t$ independently gives us the optimal calibration function $g_t$ for every time step.}
For any synthetic sample $\hat{\boldsymbol{x}}_{0}$ from DDPM, the calibration is \changed{done through the composition of two ECDFs as}
\begin{equation}
    \label{eq:calibration}
    \hat{x}_{0,t}^\prime \leftarrow F_t^{*-1}(F_t^{\prime}(\hat{x}_{0,t})) \;\; \forall t.
\end{equation}
\changed{This calibration only requires estimating one-dimensional ECDFs, eliminating the need for any additional neural networks. It happens after the training of DDPM, and the information encoded by the trained DDPM is implicitly contained within $F^\prime_t$.}

\section{Case of Study}
\label{sec:case-of-study}
\subsection{Datasets}
We show \ours's flexibility and capability of modeling high-dimensional data by selecting a diverse set of data sources across various energy domains, time resolutions, and at both customer household and transformer levels. We preprocess all datasets by splitting them into daily profiles, and we see these profiles as identical independent samples. Depending on the time resolution, the daily profile time series lengths range from $24$ at a 1-hour resolution to $1440$ at a 1-minute resolution. We categorize the datasets into three classes. First, residential electricity load profiles at the customer level. This type of data comes from the Low Carbon London (LCL)~\cite{schofield2016LowCarbonLondon} project, WPuQ~\cite{schlemminger2022DatasetElectricalSinglefamily} project and CoSSMic~\cite{amato2014SLACollaboratingSmart} project. Second, residential household heat pump electricity consumption data from the WPuQ project. Third, transformer-level electricity consumption and PV generation data from the WPuQ project. We summarize the selected datasets in Table~\ref{tab:datasets}. \changed{In our experiments, we train an independent \ours for each dataset and resolution and evaluate their performance. Notably, it is possible to train one model on multiple resolutions/datasets altogether, but this work aims at validating the effectiveness of \ours on the most fundamental setting of a single data source and resolution. }
\begin{table}[t]
    \centering
    \caption{Overview of datasets in case of study.}
    \scalebox{0.75}{
    \begin{tabular}{ccccc}
        \toprule
        Name & Country & Type & Resolution & Level \\
        \midrule
        WPuQ~\cite{schlemminger2022DatasetElectricalSinglefamily} & Germany 
        & Heat Pump & \SI{1}{\minute}$\sim$\SI{1}{\hour} & Household \\
        WPuQ~\cite{schlemminger2022DatasetElectricalSinglefamily} & Germany 
        & Electricity & \SI{1}{\minute}$\sim$\SI{1}{\hour} & Substation \\
        WPuQ~\cite{schlemminger2022DatasetElectricalSinglefamily} & Germany 
        & PV & \SI{1}{\minute}$\sim$\SI{1}{\hour} & Multi-household \\
        LCL~\cite{schofield2016LowCarbonLondon} & UK & Electricity & \SI{30}{\minute}$\sim$\SI{1}{\hour} & Household \\
        CoSSMic~\cite{amato2014SLACollaboratingSmart} & Germany & Electricity & \SI{1}{\minute}$\sim$\SI{1}{\hour} & Household \\
        CoSSMic~\cite{amato2014SLACollaboratingSmart} & Germany & PV & \SI{1}{\minute}$\sim$\SI{1}{\hour} & Household \\
        \bottomrule
    \end{tabular}}%
    \label{tab:datasets}%
    \vspace{-4mm}
\end{table}

\subsection{Evaluation Metrics}
Because of the high-dimensional nature of time series data, it is difficult to apply conventional probability theory-based measures to examine the joint distribution divergence between the real and generated samples. Nonetheless, several metrics have been established to measure the quality of synthetic data. We will give a brief introduction to these metrics, while the details and equations can be found in~\cite{xia2024ComparativeAssessmentGenerative}. 
\todo{Meanwhile, we test our generated data in downstream tasks, namely load forecast (and ...). }

\subsubsection{Gaussian Frechét Distance}
The Frechét distance (FD) was proposed and adapted to compare the similarity between two probability distributions. Despite the intractability of FD for high-dimensional joint distributions, an analytical solution exists between two multivariate Gaussian distributions. Therefore, we can generalize FD to Gaussian Frechét distance (GFD) to measure the similarity between any two multivariate joint distributions because it quantifies simultaneously the differences in the mean and covariance matrices. 

\subsubsection{Maximum Mean Discrepancy (MMD)}
Maximum Mean Discrepancy is a kernel-based disparity measure. It embeds samples of arbitrary data space (e.g., $\mathbb{R}^d$) into a reproducing kernel Hilbert space (RKHS) and compares two distributions by the largest difference in expectations over their embeddings in the RKHS. MMD measures both the dependency structure and the marginal distribution through an implicit feature mapping via the kernels.

\subsubsection{Wasserstein Distance (WD)}
Wasserstein distance measures the minimum distance between two distributions with the optimal coupling.
A coupling of $\boldsymbol{x}$ and $\boldsymbol{y}$ is a joint distribution over $\text{Concat}(\boldsymbol{x},\boldsymbol{y})$, whose marginal distributions satisfies $\int_y c(\boldsymbol{x}, \boldsymbol{y}) = p_X(\boldsymbol{x})$ and $\int_x c(\boldsymbol{x}, \boldsymbol{y}) = p_Y(\boldsymbol{y})$. Taking the infimum suggests that the coupling $c$ in Wasserstein distance seeks to connect $\boldsymbol{x}$ and $\boldsymbol{y}$ in the shortest path possible. In other words, WD measures the shortest distance between two distributions. However, it is generally intractable to find such coupling between two high-dimensional joint distributions. Therefore, we only compare the Wasserstein distance between their marginal distributions in the following section.

\subsubsection{Kullback-Leibler (KL) Divergence}
Unlike WD, KL divergence measures the discrepancy between two distributions from the perspective of the information theory. 
Unfortunately, calculating KL divergence between high-dimensional distributions is also generally not possible. Again, we will evaluate KL divergence between the marginal distributions. 

\subsubsection{Kolmogorov-Smirnov (KS) Statistic}
The two-sample KS test is a procedure to check whether two underlying one-dimensional distributions differ. It exploits the KS statistic, defined as the largest difference between two CDFs across all $x$. Similarly, we will evaluate the KS statistic between the marginal distributions, as KS statistic is not tractable for high-dimensional joint distribution. 

\changed{Quantifying the discrepancies between high-dimensional distributions remains a fundamental challenge and demands careful consideration. Among the five evaluation metrics we use, three of them, namely WD, KL, and KS, operate solely on the marginal distributions, overlooking the temporal dependency inherent in the time series. While these widely adopted metrics can effectively assess aspects like peak timing and magnitude, they fail to measure crucial temporal characteristics such as peak duration. The MMD and GFD can partially address the temporal dependencies, but each has limitations: MMD operates in an abstract RKHS that lacks interpretability, while GFD only compares the first- and second-order moments of the distributions. To complement these limitations, we employ both direct visualization and dimensionality-reduced representations of the time series data as qualitative assessment tools.}

\changed{Precisely calculating the metrics above requires taking the expectation over the whole data distribution. However, in practice, we only have a finite dataset. This often leads to statistical errors. To minimize this impact, we employ a bootstrapping strategy. Namely, we calculate the metrics using a random subset of the data and repeat $10$ times to get the average values and standard deviations.}

\subsection{Model Setup}
Due to the versatility of our framework, we use the same setup for all experiments in this section. Specifically, we implement \ours in PyTorch. The Transformer blocks have $L=12$ layers and $d^\prime=512$ neurons in each block. The number of neurons in the FFN is $4d^\prime$. For training, a learning rate of $0.0001$ is used, and we train the neural networks for \changed{$100000$} iterations with the AdamW optimizer~\cite{loshchilov2019DecoupledWeightDecay}, \changed{regardless of the training sample size}. Meanwhile, as a common practice of deep generative models, we keep an exponential moving average (EMA) version of the model weights, and we always use the EMA weights for sampling\cite{ho2020DenoisingDiffusionProbabilistic}. This can be seen as an extra measure to stabilize the training. In terms of diffusion, we use $S=4000$ diffusion steps and accelerate the sampling using DPM-Solver~\cite{lu2022DPMSolverFastODE} with 100 steps. A cosine noise schedule is adopted as in~\cite{nichol2021ImprovedDenoisingDiffusion}. We only perform minimum preprocessing on the data, i.e., all data are linearly scaled to $[-1,1]$ using the minimum and the maximum values. 
\changed{We use four baseline models, namely $10$-component GMMs, t-Copula~\cite{duque2021ConditionalMultivariateElliptical}, VAEs~\cite{kingma2014AutoEncodingVariationalBayes, xia2024ComparativeAssessmentGenerative}, and GANs~\cite{chen2018ModelfreeRenewableScenario, zheng2021GenerativeAdversarialNetworksbased}. VAE and GAN are two commonly used deep generative models. We follow a similar setup of VAE from \cite{xia2024ComparativeAssessmentGenerative}, and the architecture of a Wasserstein GAN from~\cite{chen2018ModelfreeRenewableScenario}. The encoder and decoder of VAE each have 5 fully connected layers with GELU~\cite{hendrycks2023GaussianErrorLinear} activation units. The Wasserstein GAN has a 6-layer convolutional generator and a 5-layer convolutional discriminator. We train these models similarly as \ours with a constant learning rate of $0.0001$ and $500$ epochs. 
Meanwhile, GMM and t-Copula are conventional statistical models that are designed to achieve high scores in terms of our evaluation metrics. GMM are optimized to learn the data's means and covariances, yielding low GFD and MMD scores. In contrast, t-Copula leverages the ECDF of the data, resulting in strong KL, WD, and KS scores. }

\section{Results}
\label{sec:results}

\todo{peak-time distribution etc.}

\subsection{Customer Level Evaluation}
\subsubsection{Heat Pump Consumption}
\begin{table}[t]
    \centering
    \caption{Evaluation Metric Results on WPuQ. Best results are \underline{underlined}. }
    \scalebox{0.75}{
    \changed{\begin{tabular}{lcrrrrr}
        \toprule
        Model & Res. & MMD & GFD & KL & WD & KS \\
        \midrule
        GMM & \multirow{12}{*}{\SI{1}{\minute}} & $\underline{0.0008}$ & $6.4440$ & $1.1469$ & $0.0401$ & $0.3327$ \\
         &  & $\pm~0.0001$ & $\pm~0.2265$ & $\pm~0.1437$ & $\pm~0.0004$ & $\pm~0.0015$ \\
        \changedii{t-Copula$^*$} &  & \changedii{$0.0014$} & \changedii{$\underline{5.5552}$} & \changedii{$0.0041$} & \changedii{$0.0042$} & \changedii{$\underline{0.0644}$} \\
         &  & \changedii{$\pm~0.0001$} & \changedii{$\pm~0.1503$} & \changedii{$\pm~0.0006$} & \changedii{$\pm~0.0005$} & \changedii{$\pm~0.0036$} \\
        VAE &  & $0.0260$ & $18.8498$ & $1.2489$ & $0.0385$ & $0.4685$ \\
         &  & $\pm~0.0007$ & $\pm~0.5886$ & $\pm~0.0947$ & $\pm~0.0007$ & $\pm~0.0101$ \\
        GAN &  & $0.4935$ & $68.4609$ & $0.5623$ & $0.0710$ & $0.3510$ \\
         &  & $\pm~0.0041$ & $\pm~0.4728$ & $\pm~0.0103$ & $\pm~0.0004$ & $\pm~0.0012$ \\
        \oursabbr &  & $0.0021$ & $6.5684$ & $0.0811$ & $0.0058$ & $0.2940$ \\
         &  & $\pm~0.0002$ & $\pm~0.2202$ & $\pm~0.0049$ & $\pm~0.0005$ & $\pm~0.0035$ \\
        \oursabbrcal &  & $0.0009$ & $7.0441$ & $\underline{0.0014}$ & $\underline{0.0019}$ & $\underline{0.2210}$ \\
         &  & $\pm~0.0002$ & $\pm~0.1933$ & $\pm~0.0004$ & $\pm~0.0008$ & $\pm~0.0030$ \\
        \midrule
        GMM & \multirow{12}{*}{\SI{15}{\minute}} & $0.0008$ & $\underline{0.0719}$ & $0.2107$ & $0.0234$ & $0.2501$ \\
         &  & $\pm~0.0002$ & $\pm~0.0029$ & $\pm~0.0341$ & $\pm~0.0009$ & $\pm~0.0039$ \\
        t-Copula &  & $0.0022$ & $0.1667$ & $0.0014$ & $0.0022$ & $\underline{0.0172}$ \\
         &  & $\pm~0.0004$ & $\pm~0.0085$ & $\pm~0.0003$ & $\pm~0.0010$ & $\pm~0.0039$ \\
        VAE &  & $0.0274$ & $0.7846$ & $0.9345$ & $0.0351$ & $0.5148$ \\
         &  & $\pm~0.0007$ & $\pm~0.0184$ & $\pm~0.0327$ & $\pm~0.0008$ & $\pm~0.0043$ \\
        GAN &  & $0.0234$ & $0.5620$ & $0.1114$ & $0.0232$ & $0.2543$ \\
         &  & $\pm~0.0023$ & $\pm~0.0290$ & $\pm~0.0086$ & $\pm~0.0020$ & $\pm~0.0042$ \\
        \oursabbr &  & $0.0033$ & $0.0760$ & $0.1013$ & $0.0110$ & $0.1929$ \\
         &  & $\pm~0.0002$ & $\pm~0.0022$ & $\pm~0.0027$ & $\pm~0.0009$ & $\pm~0.0031$ \\
        \oursabbrcal &  & $\underline{0.0006}$ & $0.0727$ & $\underline{0.0013}$ & $\underline{0.0021}$ & $0.1030$ \\
         &  & $\pm~0.0001$ & $\pm~0.0022$ & $\pm~0.0003$ & $\pm~0.0009$ & $\pm~0.0012$ \\
        \midrule
        GMM & \multirow{12}{*}{\SI{30}{\minute}} & $0.0007$ & $0.0138$ & $0.0932$ & $0.0164$ & $0.1881$ \\
         &  & $\pm~0.0001$ & $\pm~0.0010$ & $\pm~0.0397$ & $\pm~0.0010$ & $\pm~0.0041$ \\
        t-Copula &  & $0.0014$ & $0.0426$ & $0.0017$ & $0.0027$ & $\underline{0.0138}$ \\
         &  & $\pm~0.0002$ & $\pm~0.0047$ & $\pm~0.0003$ & $\pm~0.0010$ & $\pm~0.0035$ \\
        VAE &  & $0.0244$ & $0.3078$ & $0.8029$ & $0.0343$ & $0.4748$ \\
         &  & $\pm~0.0009$ & $\pm~0.0104$ & $\pm~0.0155$ & $\pm~0.0010$ & $\pm~0.0036$ \\
        GAN &  & $0.0117$ & $0.1371$ & $0.0439$ & $0.0185$ & $0.1324$ \\
         &  & $\pm~0.0009$ & $\pm~0.0084$ & $\pm~0.0022$ & $\pm~0.0018$ & $\pm~0.0030$ \\
        \oursabbr &  & $0.0019$ & $0.0162$ & $0.0443$ & $0.0086$ & $0.1777$ \\
         &  & $\pm~0.0002$ & $\pm~0.0010$ & $\pm~0.0026$ & $\pm~0.0009$ & $\pm~0.0033$ \\
        \oursabbrcal &  & $\underline{0.0006}$ & $\underline{0.0136}$ & $\underline{0.0016}$ & $\underline{0.0019}$ & $0.1039$ \\
         &  & $\pm~0.0000$ & $\pm~0.0015$ & $\pm~0.0003$ & $\pm~0.0009$ & $\pm~0.0022$ \\
        \midrule
        GMM & \multirow{12}{*}{\SI{1}{\hour}} & $0.0009$ & $\underline{0.0020}$ & $0.0696$ & $0.0111$ & $0.1596$ \\
         &  & $\pm~0.0002$ & $\pm~0.0004$ & $\pm~0.0350$ & $\pm~0.0010$ & $\pm~0.0037$ \\
        t-Copula &  & $0.0009$ & $0.0074$ & $0.0026$ & $\underline{0.0026}$ & $\underline{0.0141}$ \\
         &  & $\pm~0.0001$ & $\pm~0.0013$ & $\pm~0.0006$ & $\pm~0.0014$ & $\pm~0.0055$ \\
        VAE &  & $0.0200$ & $0.0901$ & $0.6434$ & $0.0287$ & $0.4164$ \\
         &  & $\pm~0.0010$ & $\pm~0.0052$ & $\pm~0.0116$ & $\pm~0.0014$ & $\pm~0.0031$ \\
        GAN &  & $0.0120$ & $0.1950$ & $0.0508$ & $0.0136$ & $0.1497$ \\
         &  & $\pm~0.0004$ & $\pm~0.0096$ & $\pm~0.0031$ & $\pm~0.0010$ & $\pm~0.0029$ \\
        \oursabbr &  & $0.0014$ & $0.0034$ & $0.0306$ & $0.0071$ & $0.1712$ \\
         &  & $\pm~0.0002$ & $\pm~0.0005$ & $\pm~0.0021$ & $\pm~0.0010$ & $\pm~0.0020$ \\
        \oursabbrcal &  & $\underline{0.0006}$ & $0.0026$ & $\underline{0.0020}$ & $0.0027$ & $0.1016$ \\
         &  & $\pm~0.0001$ & $\pm~0.0006$ & $\pm~0.0005$ & $\pm~0.0014$ & $\pm~0.0025$ \\
        \bottomrule
        \multicolumn{7}{l}{The values following $\pm$ are standard deviations.} \\
        \multicolumn{7}{l}{\changedii{$^*$ From repeated retrying with different random seeds for over 10 hours.}} \\
        \multicolumn{7}{l}{\oursabbr: the proposed \ours framework.} \\
        \multicolumn{7}{l}{\oursabbrcal: the proposed \ours framework with Marginal Calibration.}
    \end{tabular}}}%
    \label{tab:wpuq-perf}%
    \vspace{-4mm}
\end{table}

\begin{figure}[t]
    \centering
    \includegraphics[width=0.85\linewidth]{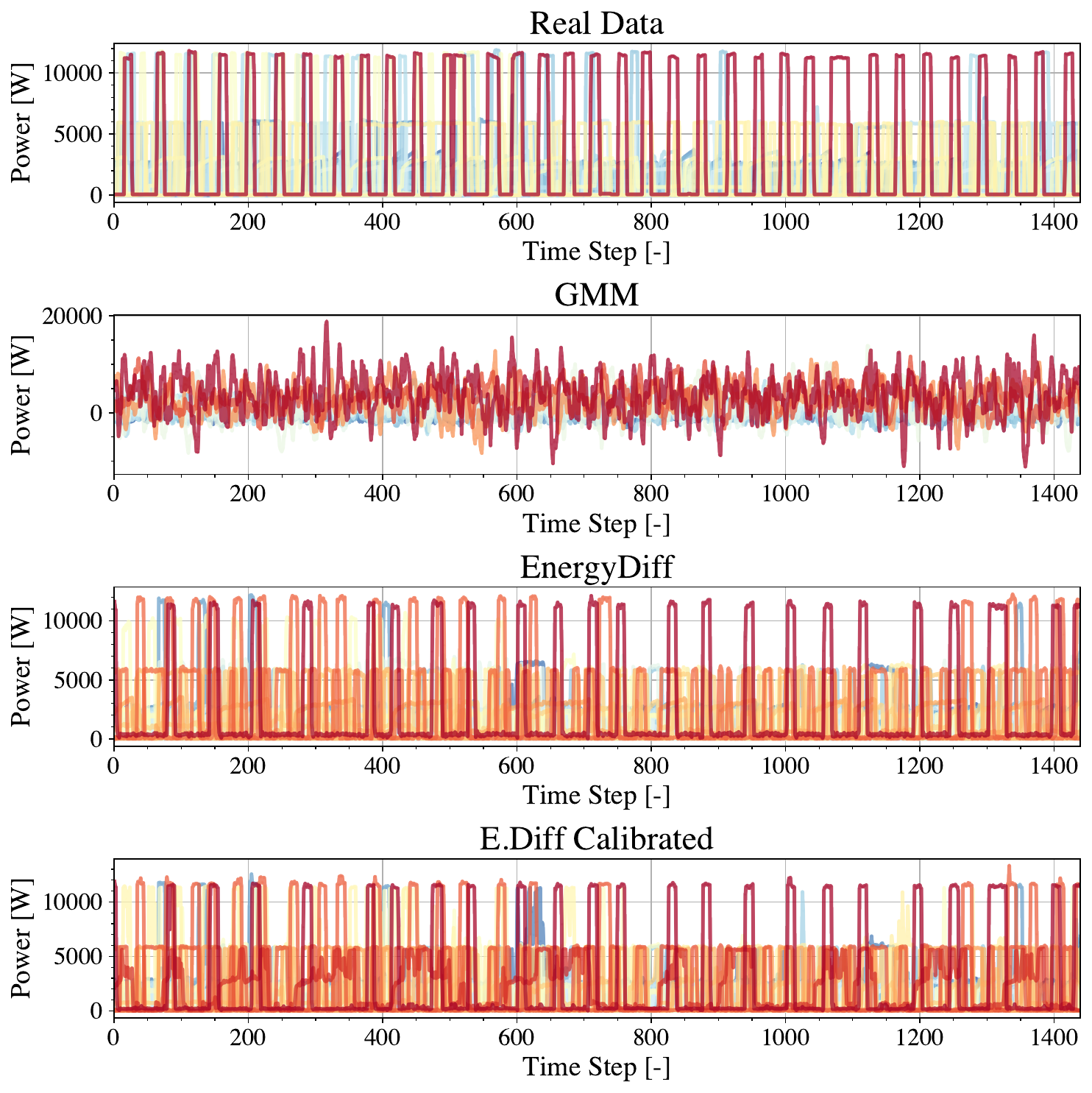}
    \caption{\changed{WPuQ heat pump consumption curves at 1-minute resolution. \changedii{The color of each curve represents the total sum over the day. Red suggests a higher consumption and blue suggests lower.} Our model captures the periodical square wave-like pattern while GMM does not. Note that for visibility, only 100 randomly selected samples are plotted in this figure. They only represent a small fraction of the real/synthetic data.}}
    \label{fig:samples-hp-1min}%
    \vspace{-2mm}
\end{figure}

For heat pump data \changed{across all resolutions, \ours with calibration consistently achieve either the best score or close (within one standard deviation range) to best scores\changedii{, as shown in Table~\ref{tab:wpuq-perf}}. The only exceptions are 1) at 1-minute resolution, \changed{t-Copula and GMM achieve} a better GFD, \changedii{2) t-Copula had best KS scores,} 3) and at 15-minute and 1-hour resolutions, t-Copula shows the lowest KS score. However, even in these two cases, GMM and t-Copula do not overperform \ours completely. GMM has a KL score of $1.1449$ $vs.$ $0.0811$ by \ours at 1 minute, while t-Copula has two times higher GFD than \ours at 15 minutes and 1 hour.} 

\changedii{Despite of t-Copula's competitiveness, t-Copula's fitting algorithm suffers from severe numeric instability. The convergence of the algorithm is sensitive to initial values, i.e., the random seeds used. This issue is not severe when the data dimension is low, for example, from the 15-minute to 1-hour resolutions. But it is becomes crucial at the 1-minute resolution when the training set is large (over a thousand samples). More than ten hours were needed to get the results presented in Table~\ref{tab:wpuq-perf}. This is not realistic in practice. We discuss the convergence and computation complexity in Section~\ref{sec:computation-time}.}

\changed{Overall, the deep learning baseline models, VAE and GAN, are not as competitive as conventional models, GMM and t-Copula. VAE's strength lies in compressing data by representing them in a low-dimensional latent space and reconstructing them with little loss~\cite{peebles2023ScalableDiffusionModels}. However, it has limited capability of generating data. This is because generating data with VAE relies on the assumption that the data has an isotropic Gaussian distribution in the latent space. In reality, this is not the case. It is required to train another model to learn the latent distribution in order to generate good-quality data~\cite{chai2024FaradaySyntheticSmart, peebles2023ScalableDiffusionModels}. On the other hand, reaching a GFD of $68.4609$ at the 1-minute case suggests GAN is unstable at a high resolution. In contrast, despite having a similar training scheme, \ours achieves a stable score across all resolutions, showing no sign of collapsing or divergence.} 

\changed{From 1-hour to 1-minute resolution, the metric values of all models (i.e., GMM, t-Copula, VAEs, GANs, and \ours) have a strong trend of deterioration as the resolution grows, suggesting the difficulty is increasing.} At the 1-hour resolution, \ours and GMM, can achieve $1\times 10^{-3}$ level of GFD, but this number increases significantly \changed{by three orders of magnitude} at 1-minute resolution. 
\changed{This gap showcases the challenge of high-resolution data modeling. As we will show later in Sec.~\ref{sec:computation-time}, the 1-minute scenarios are exactly where t-Copula and GAN models fail the most. At this resolution, GMM has a large KL, t-Copula did not converge, VAE and GAN has very large GFD scores. The only model that performs well consistently across all metrics is the proposed \ours. }

We present randomly \changed{and independently} selected $100$ real and generated heat pump consumption data samples in Fig.~\ref{fig:samples-hp-1min}, corresponding to the numerical results in Table~\ref{tab:wpuq-perf}. \changed{We use Fig.~\ref{fig:samples-hp-1min} as a supporting intuitive illustration of the visible patterns of the real and generated data, but not a rigorous comparison.} Despite the MMD between GMM and \ours being similar, \changed{and GMM having better GFD}, we can observe in Fig.~\ref{fig:samples-hp-1min} that the synthetic samples of GMM are clearly unrealistic, as they do not contain the periodical temporal patterns of the real data. 
\changed{While the real data has a minimum of \SI{0}{\watt}}, GMM produces unrealistic peaks of over \kw{20} and negative values below \kw{-10}. On the other hand, \ours successfully captures both the periodicity pattern and the value range. \changed{We highlight that despite the numerical improvements brought by marginal calibration, it does not cause significant changes in the shape of the profiles. This is expected when the uncalibrated model already learns the data distribution well. It also shows that our calibration technique is indeed minimizing the changes on the data while satisfying the constraints in \eqref{eq:calibration}.}

\begin{figure*}[t]
    \centering
    \includegraphics[width=0.95\linewidth]{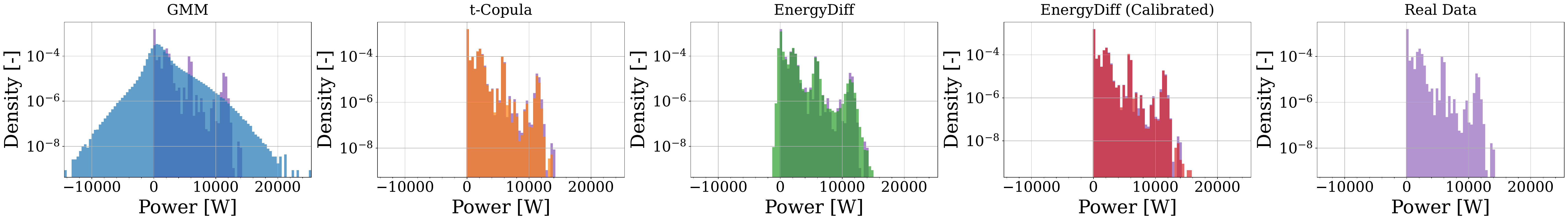}
    \caption{Histograms of WPuQ heat pump consumption at 1 minute resolution. GMM fails to find the correct range of data. t-Copula fits exactly the empirical CDF. The proposed \ours fits the smoothed CDF, while after calibration, \ours fits the CDF perfectly.}
    \label{fig:hist-hp-1min}
\end{figure*}

Furthermore, in Fig.~\ref{fig:hist-hp-1min}, we present the histograms of the same heat pump consumption data at the 1-minute resolution. To get the histogram, we count the consumption power of all time steps. The t-Copula model closely matches the histogram because it is designed to do so. \ours without calibration can capture the general pattern of the histogram but appears more smoothed, while the calibrated data matches the histogram exactly. In contrast, the GMM model fails to capture the distribution, as it struggles to find the correct support for the distribution and fails to capture the different modes of distribution. \changed{We will delve into a more detailed discussion about marginal calibration in Sec.~\ref{sec:marginal-calibration}.}

\subsubsection{Residential Electricity Load Profile}
We evaluate \ours's capability to model electricity load profile data on two datasets, LCL at 30-minute and 1-hour resolutions and CoSSMic at 1-minute resolutions. For the LCL dataset, as shown in Table~\ref{tab:lcl-rlp-perf}, our model and GMM achieve similarly good results in MMD and GFD at both time resolutions \changed{and \ours with calibration performs better in KL and WD, while t-Copula performs worse but still remains competitive.} Turning to the CoSSMic Dataset \changedii{in Table~\ref{tab:cossmic-grid-import-residential-perf}}, our model delivers best or close to best results across all metrics. \changed{GMM's GFD score exceeds ours by a small margin, but our model achieves better scores of all the other metrics.} \mbox{t-Copula}, \changed{at the 1-minute resolution}, \changed{again}, fails to converge during the fitting process, \changed{but achieves competitive scores across most metrics at all the other resolutions}. As evidenced by the numerical results on these two datasets, \ours demonstrate strong performance in generating residential electricity load profiles across various time resolutions. 

\begin{table}[t]
    \centering
    \caption{Evaluation Metric Results on LCL Residential Electricity Load Profiles}
    \scalebox{0.75}{
    \changed{\begin{tabular}{lcrrrrr}
        \toprule
        Model & Res. & MMD & GFD & KL & WD & KS \\
        \midrule
        GMM & \multirow{12}{*}{\SI{30}{\minute}} & $\underline{0.0016}$ & $0.0167$ & $0.0522$ & $0.0082$ & $0.1079$ \\
         &  & $\pm~0.0003$ & $\pm~0.0023$ & $\pm~0.0117$ & $\pm~0.0006$ & $\pm~0.0021$ \\
        t-Copula &  & $0.0029$ & $0.0169$ & $0.0235$ & $0.0039$ & $0.0803$ \\
         &  & $\pm~0.0004$ & $\pm~0.0012$ & $\pm~0.0020$ & $\pm~0.0003$ & $\pm~0.0042$ \\
        VAE &  & $0.1235$ & $0.1421$ & $0.3623$ & $0.0289$ & $0.6069$ \\
         &  & $\pm~0.0047$ & $\pm~0.0041$ & $\pm~0.1290$ & $\pm~0.0001$ & $\pm~0.0020$ \\
        GAN &  & $0.0707$ & $0.1069$ & $0.5953$ & $0.0220$ & $0.4690$ \\
         &  & $\pm~0.0025$ & $\pm~0.0062$ & $\pm~0.0326$ & $\pm~0.0005$ & $\pm~0.0022$ \\
        \oursabbr &  & $0.0020$ & $0.0141$ & $0.0282$ & $0.0032$ & $0.0842$ \\
         &  & $\pm~0.0002$ & $\pm~0.0029$ & $\pm~0.0023$ & $\pm~0.0003$ & $\pm~0.0017$ \\
        \oursabbrcal &  & $0.0017$ & $\underline{0.0134}$ & $\underline{0.0044}$ & $\underline{0.0022}$ & $\underline{0.0745}$ \\
         &  & $\pm~0.0002$ & $\pm~0.0039$ & $\pm~0.0008$ & $\pm~0.0004$ & $\pm~0.0020$ \\
        \midrule
        GMM & \multirow{12}{*}{\SI{1}{\hour}} & $\underline{0.0015}$ & $\underline{0.0140}$ & $0.0480$ & $0.0074$ & $0.1081$ \\
         &  & $\pm~0.0002$ & $\pm~0.0022$ & $\pm~0.0116$ & $\pm~0.0003$ & $\pm~0.0024$ \\
        t-Copula &  & $0.0039$ & $0.0191$ & $0.0334$ & $0.0042$ & $0.0962$ \\
         &  & $\pm~0.0004$ & $\pm~0.0029$ & $\pm~0.0027$ & $\pm~0.0003$ & $\pm~0.0051$ \\
        VAE &  & $0.0686$ & $0.1453$ & $0.3793$ & $0.0232$ & $0.4625$ \\
         &  & $\pm~0.0018$ & $\pm~0.0056$ & $\pm~0.1340$ & $\pm~0.0004$ & $\pm~0.0031$ \\
        GAN &  & $0.0565$ & $0.1052$ & $0.4572$ & $0.0201$ & $0.3608$ \\
         &  & $\pm~0.0020$ & $\pm~0.0046$ & $\pm~0.0123$ & $\pm~0.0005$ & $\pm~0.0023$ \\
        \oursabbr &  & $0.0021$ & $0.0148$ & $0.0332$ & $0.0037$ & $0.0789$ \\
         &  & $\pm~0.0002$ & $\pm~0.0022$ & $\pm~0.0022$ & $\pm~0.0004$ & $\pm~0.0028$ \\
        \oursabbrcal &  & $0.0018$ & $0.0164$ & $\underline{0.0048}$ & $\underline{0.0021}$ & $\underline{0.0694}$ \\
         &  & $\pm~0.0002$ & $\pm~0.0023$ & $\pm~0.0009$ & $\pm~0.0003$ & $\pm~0.0014$ \\
        \bottomrule
    \end{tabular}}}%
    \label{tab:lcl-rlp-perf}%
    \vspace{-4mm}
\end{table}

\begin{table}[t]
    \centering
    \caption{Evaluation Metric Results on CoSSMic Residential Electricity Load Profiles}
    \scalebox{0.75}{
    \changed{\begin{tabular}{lcrrrrr}
        \toprule
        Model & Res. & MMD & GFD & KL & WD & KS \\
        \midrule
        GMM & \multirow{12}{*}{\SI{1}{\minute}} & $0.0032$ & $\underline{4.0624}$ & $0.7387$ & $0.0299$ & $0.2034$ \\
         &  & $\pm~0.0004$ & $\pm~0.1096$ & $\pm~0.0771$ & $\pm~0.0005$ & $\pm~0.0018$ \\
        t-Copula &
                    & \multirow{2}{*}{\tripleast} & \multirow{2}{*}{\tripleast} & \multirow{2}{*}{\tripleast} & \multirow{2}{*}{\tripleast} & \multirow{2}{*}{\tripleast} \\
        &&&&&& \\
        VAE &  & $0.0810$ & $10.3594$ & $0.9515$ & $0.0378$ & $0.2884$ \\
         &  & $\pm~0.0021$ & $\pm~0.2089$ & $\pm~0.0227$ & $\pm~0.0004$ & $\pm~0.0042$ \\
        GAN &  & $0.0729$ & $14.3728$ & $0.5681$ & $0.0420$ & $0.2884$ \\
         &  & $\pm~0.0057$ & $\pm~0.1089$ & $\pm~0.0105$ & $\pm~0.0007$ & $\pm~0.0059$ \\
        \oursabbr &  & $0.0141$ & $5.1522$ & $0.5834$ & $0.0153$ & $0.2086$ \\
         &  & $\pm~0.0007$ & $\pm~0.1887$ & $\pm~0.0104$ & $\pm~0.0005$ & $\pm~0.0049$ \\
        \oursabbrcal &  & $\underline{0.0032}$ & $5.1848$ & $\underline{0.0029}$ & $\underline{0.0038}$ & $\underline{0.0197}$ \\
         &  & $\pm~0.0005$ & $\pm~0.0997$ & $\pm~0.0007$ & $\pm~0.0009$ & $\pm~0.0047$ \\
        \midrule
        GMM & \multirow{12}{*}{\SI{15}{\minute}} & $\underline{0.0026}$ & $\underline{0.0821}$ & $0.0819$ & $0.0238$ & $0.1398$ \\
         &  & $\pm~0.0005$ & $\pm~0.0050$ & $\pm~0.0081$ & $\pm~0.0013$ & $\pm~0.0020$ \\
        t-Copula &  & $0.0786$ & $0.9448$ & $0.3040$ & $0.0801$ & $0.3159$ \\
         &  & $\pm~0.0039$ & $\pm~0.0306$ & $\pm~0.0097$ & $\pm~0.0014$ & $\pm~0.0074$ \\
        VAE &  & $0.0453$ & $0.7397$ & $0.2333$ & $0.0345$ & $0.2263$ \\
         &  & $\pm~0.0015$ & $\pm~0.0219$ & $\pm~0.0097$ & $\pm~0.0009$ & $\pm~0.0060$ \\
        GAN &  & $0.0993$ & $0.9066$ & $0.2944$ & $0.0592$ & $0.3588$ \\
         &  & $\pm~0.0037$ & $\pm~0.0478$ & $\pm~0.0136$ & $\pm~0.0019$ & $\pm~0.0039$ \\
        \oursabbr &  & $0.0078$ & $0.1100$ & $0.0462$ & $0.0142$ & $0.0974$ \\
         &  & $\pm~0.0008$ & $\pm~0.0064$ & $\pm~0.0031$ & $\pm~0.0008$ & $\pm~0.0074$ \\
        \oursabbrcal &  & $0.0032$ & $0.0846$ & $\underline{0.0063}$ & $\underline{0.0057}$ & $\underline{0.0261}$ \\
         &  & $\pm~0.0005$ & $\pm~0.0039$ & $\pm~0.0018$ & $\pm~0.0014$ & $\pm~0.0049$ \\
        \midrule
        GMM & \multirow{12}{*}{\SI{30}{\minute}} & $0.0041$ & $0.0263$ & $0.0633$ & $0.0201$ & $0.1137$ \\
         &  & $\pm~0.0008$ & $\pm~0.0037$ & $\pm~0.0150$ & $\pm~0.0015$ & $\pm~0.0021$ \\
        t-Copula &  & $0.0545$ & $0.2758$ & $0.2003$ & $0.0637$ & $0.2491$ \\
         &  & $\pm~0.0057$ & $\pm~0.0184$ & $\pm~0.0142$ & $\pm~0.0027$ & $\pm~0.0096$ \\
        VAE &  & $0.0384$ & $0.3494$ & $0.1720$ & $0.0296$ & $0.1786$ \\
         &  & $\pm~0.0010$ & $\pm~0.0112$ & $\pm~0.0072$ & $\pm~0.0012$ & $\pm~0.0069$ \\
        GAN &  & $0.1167$ & $0.6197$ & $0.2569$ & $0.0655$ & $0.2770$ \\
         &  & $\pm~0.0028$ & $\pm~0.0141$ & $\pm~0.0087$ & $\pm~0.0011$ & $\pm~0.0046$ \\
        \oursabbr &  & $\underline{0.0028}$ & $\underline{0.0217}$ & $0.0332$ & $\underline{0.0065}$ & $0.0396$ \\
         &  & $\pm~0.0003$ & $\pm~0.0020$ & $\pm~0.0021$ & $\pm~0.0007$ & $\pm~0.0037$ \\
        \oursabbrcal &  & $0.0032$ & $0.0240$ & $\underline{0.0079}$ & $0.0070$ & $\underline{0.0288}$ \\
         &  & $\pm~0.0005$ & $\pm~0.0028$ & $\pm~0.0017$ & $\pm~0.0019$ & $\pm~0.0050$ \\
        \midrule
        GMM & \multirow{12}{*}{\SI{1}{\hour}} & $0.0040$ & $0.0110$ & $0.0496$ & $0.0182$ & $0.0913$ \\
         &  & $\pm~0.0008$ & $\pm~0.0016$ & $\pm~0.0079$ & $\pm~0.0018$ & $\pm~0.0025$ \\
        t-Copula &  & $0.0287$ & $0.0953$ & $0.1074$ & $0.0528$ & $0.1652$ \\
         &  & $\pm~0.0019$ & $\pm~0.0114$ & $\pm~0.0065$ & $\pm~0.0032$ & $\pm~0.0067$ \\
        VAE &  & $0.0207$ & $0.1456$ & $0.1020$ & $0.0246$ & $0.0964$ \\
         &  & $\pm~0.0009$ & $\pm~0.0069$ & $\pm~0.0075$ & $\pm~0.0011$ & $\pm~0.0057$ \\
        GAN &  & $0.0451$ & $0.2844$ & $0.1743$ & $0.0370$ & $0.1670$ \\
         &  & $\pm~0.0018$ & $\pm~0.0113$ & $\pm~0.0059$ & $\pm~0.0028$ & $\pm~0.0059$ \\
        \oursabbr &  & $\underline{0.0025}$ & $\underline{0.0106}$ & $0.0254$ & $\underline{0.0069}$ & $0.0335$ \\
         &  & $\pm~0.0003$ & $\pm~0.0008$ & $\pm~0.0017$ & $\pm~0.0017$ & $\pm~0.0050$ \\
        \oursabbrcal &  & $0.0029$ & $0.0117$ & $\underline{0.0088}$ & $0.0088$ & $\underline{0.0245}$ \\
         &  & $\pm~0.0006$ & $\pm~0.0019$ & $\pm~0.0018$ & $\pm~0.0016$ & $\pm~0.0039$ \\
        \bottomrule
        \multicolumn{7}{l}{\tripleast indicates the model did not converge.}
    \end{tabular}}}%
    \label{tab:cossmic-grid-import-residential-perf}%
    \vspace{-4mm}
\end{table}

\subsubsection{Residential PV Generation}
In Table~\ref{tab:cossmic-pv-residential-perf}, we show the numerical results of the CoSSMic residential PV generation time series data. The t-Copula model failed to converge \changed{completely}. Therefore, we only compare GMM and our model. At the 1-minute resolution, our model achieved the best results in three out of the five metrics: \changed{KL, WD, and KS}, while GMM achieves better MMD and GFD, \changed{despite this gap being closed largely by marginal calibration}. \changed{This gap in MMD and GFD} is likely due to this dataset's small sample size, as only $408$ samples are available for training and $416$ samples for evaluation. \changed{At all the other resolutions}, our model \changed{achieves best to close to best scores, largely thanks to the margincal calibration}. This \changed{signals a potential weakness of \ours when the training sample size is extremely small}. \changed{In those cases, it is necessary to adopt the marginal calibration and the user should consider compensating it with} pre-training on a larger similar dataset and fine-tuning on the target dataset, as is done by other generative model research such as~\cite{peebles2023ScalableDiffusionModels}. 

\begin{table}[t]
    \centering
    \caption{Evaluation Metric Results on CoSSMic Residential PV Generation}
    \scalebox{0.75}{
    \changed{\begin{tabular}{lcrrrrr}
        \toprule
        Model & Res. & MMD & GFD & KL & WD & KS \\
        \midrule
        GMM & \multirow{12}{*}{\SI{1}{\minute}} & $\underline{0.0021}$ & $\underline{3.2551}$ & $0.2011$ & $0.0078$ & $0.3012$ \\
         &  & $\pm~0.0004$ & $\pm~0.1445$ & $\pm~0.0815$ & $\pm~0.0009$ & $\pm~0.0016$ \\
        t-Copula &
                    & \multirow{2}{*}{\tripleast} & \multirow{2}{*}{\tripleast} & \multirow{2}{*}{\tripleast} & \multirow{2}{*}{\tripleast} & \multirow{2}{*}{\tripleast} \\
        &&&&&& \\
        VAE &  & $0.0236$ & $15.2792$ & $1.5724$ & $0.0231$ & $0.3538$ \\
         &  & $\pm~0.0009$ & $\pm~0.2759$ & $\pm~0.3661$ & $\pm~0.0007$ & $\pm~0.0027$ \\
        GAN &  & $0.1445$ & $35.0172$ & $0.8735$ & $0.0768$ & $0.5836$ \\
         &  & $\pm~0.0101$ & $\pm~0.6530$ & $\pm~0.0089$ & $\pm~0.0023$ & $\pm~0.0037$ \\
        \oursabbr &  & $0.0197$ & $10.5760$ & $0.3567$ & $0.0267$ & $0.3966$ \\
         &  & $\pm~0.0030$ & $\pm~0.7083$ & $\pm~0.0097$ & $\pm~0.0022$ & $\pm~0.0025$ \\
        \oursabbrcal &  & $0.0050$ & $7.0128$ & $\underline{0.0009}$ & $\underline{0.0032}$ & $\underline{0.0059}$ \\
         &  & $\pm~0.0012$ & $\pm~0.2779$ & $\pm~0.0003$ & $\pm~0.0027$ & $\pm~0.0031$ \\
        \midrule
        GMM & \multirow{12}{*}{\SI{15}{\minute}} & $\underline{0.0020}$ & $\underline{0.0017}$ & $0.0179$ & $0.0009$ & $0.3070$ \\
         &  & $\pm~0.0003$ & $\pm~0.0014$ & $\pm~0.0397$ & $\pm~0.0001$ & $\pm~0.0013$ \\
        t-Copula &
                    & \multirow{2}{*}{\tripleast} & \multirow{2}{*}{\tripleast} & \multirow{2}{*}{\tripleast} & \multirow{2}{*}{\tripleast} & \multirow{2}{*}{\tripleast} \\
        &&&&&& \\
        VAE &  & $0.1228$ & $0.0136$ & $0.4629$ & $0.0047$ & $0.5336$ \\
         &  & $\pm~0.0405$ & $\pm~0.0033$ & $\pm~0.2468$ & $\pm~0.0001$ & $\pm~0.0003$ \\
        GAN &  & $0.2487$ & $0.0151$ & $0.2078$ & $0.0049$ & $0.5386$ \\
         &  & $\pm~0.0294$ & $\pm~0.0030$ & $\pm~0.1042$ & $\pm~0.0001$ & $\pm~0.0025$ \\
        \oursabbr &  & $0.0852$ & $0.0083$ & $0.1171$ & $0.0041$ & $0.3252$ \\
         &  & $\pm~0.0120$ & $\pm~0.0027$ & $\pm~0.0052$ & $\pm~0.0001$ & $\pm~0.0035$ \\
        \oursabbrcal &  & $0.0038$ & $0.0035$ & $\underline{0.0003}$ & $\underline{0.0002}$ & $\underline{0.0160}$ \\
         &  & $\pm~0.0008$ & $\pm~0.0030$ & $\pm~0.0002$ & $\pm~0.0001$ & $\pm~0.0034$ \\
        \midrule
        GMM & \multirow{12}{*}{\SI{30}{\minute}} & $\underline{0.0020}$ & $0.0035$ & $0.0115$ & $0.0010$ & $0.2902$ \\
         &  & $\pm~0.0003$ & $\pm~0.0044$ & $\pm~0.0232$ & $\pm~0.0001$ & $\pm~0.0017$ \\
        t-Copula &
                    & \multirow{2}{*}{\tripleast} & \multirow{2}{*}{\tripleast} & \multirow{2}{*}{\tripleast} & \multirow{2}{*}{\tripleast} & \multirow{2}{*}{\tripleast} \\
        &&&&&& \\
        VAE &  & $0.1547$ & $0.0171$ & $0.5246$ & $0.0073$ & $0.4998$ \\
         &  & $\pm~0.0304$ & $\pm~0.0029$ & $\pm~0.3045$ & $\pm~0.0001$ & $\pm~0.0005$ \\
        GAN &  & $0.2543$ & $0.0267$ & $0.2003$ & $0.0079$ & $0.5281$ \\
         &  & $\pm~0.0287$ & $\pm~0.0042$ & $\pm~0.0275$ & $\pm~0.0002$ & $\pm~0.0038$ \\
        \oursabbr &  & $0.0190$ & $0.0055$ & $0.0842$ & $0.0032$ & $0.2660$ \\
         &  & $\pm~0.0023$ & $\pm~0.0035$ & $\pm~0.0637$ & $\pm~0.0002$ & $\pm~0.0042$ \\
        \oursabbrcal &  & $0.0023$ & $\underline{0.0014}$ & $\underline{0.0005}$ & $\underline{0.0004}$ & $\underline{0.0164}$ \\
         &  & $\pm~0.0005$ & $\pm~0.0011$ & $\pm~0.0002$ & $\pm~0.0002$ & $\pm~0.0031$ \\
        \midrule
        GMM & \multirow{12}{*}{\SI{1}{\hour}} & $\underline{0.0021}$ & $0.0046$ & $0.0049$ & $0.0016$ & $0.2779$ \\
         &  & $\pm~0.0004$ & $\pm~0.0047$ & $\pm~0.0084$ & $\pm~0.0003$ & $\pm~0.0017$ \\
        t-Copula &
                    & \multirow{2}{*}{\tripleast} & \multirow{2}{*}{\tripleast} & \multirow{2}{*}{\tripleast} & \multirow{2}{*}{\tripleast} & \multirow{2}{*}{\tripleast} \\
        &&&&&& \\
        VAE &  & $0.1646$ & $0.0321$ & $0.4702$ & $0.0142$ & $0.4536$ \\
         &  & $\pm~0.0184$ & $\pm~0.0026$ & $\pm~0.2494$ & $\pm~0.0002$ & $\pm~0.0005$ \\
        GAN &  & $0.6607$ & $0.2381$ & $0.1692$ & $0.0523$ & $0.5048$ \\
         &  & $\pm~0.0175$ & $\pm~0.0035$ & $\pm~0.0523$ & $\pm~0.0005$ & $\pm~0.0025$ \\
        \oursabbr &  & $0.0040$ & $0.0026$ & $0.0193$ & $0.0023$ & $0.2157$ \\
         &  & $\pm~0.0003$ & $\pm~0.0021$ & $\pm~0.0068$ & $\pm~0.0002$ & $\pm~0.0037$ \\
        \oursabbrcal &  & $0.0023$ & $\underline{0.0023}$ & $\underline{0.0009}$ & $\underline{0.0008}$ & $\underline{0.0213}$ \\
         &  & $\pm~0.0007$ & $\pm~0.0016$ & $\pm~0.0003$ & $\pm~0.0004$ & $\pm~0.0029$ \\
        \bottomrule
    \end{tabular}}}%
    \label{tab:cossmic-pv-residential-perf}%
    \vspace{-1mm}
\end{table}

\subsection{Transformer Level Evaluation}
Our model can generate high-resolution energy time series data not only at the customer level but also at the transformer level. We present our experiment results in the WPuQ transformer electricity consumption dataset and the WPuQ transformer PV generation dataset. Both datasets have over $1500$ samples for training and evaluation. Tables~\ref{tab:wpuq-trafo-perf} and \ref{tab:wpuq-pv-perf} summarize the numerical results on these two datasets. 

\subsubsection{Residential Electricity Load Profile}
In the WPuQ transformer case, the proposed \ours exhibits superior performance in terms of all five metrics \changed{at 1-minute, 15-minute, and 1-hour resolution, as shown in Table~\ref{tab:wpuq-trafo-perf}. At the 30-minute resolution, GMM has exceeded our model in all metrics. It warns us that \ours is not perfect. However, the gap here is small; \ours still maintains its universal applicability.}

\begin{figure}[!t]
    \centering
    \includegraphics[width=0.9\linewidth]{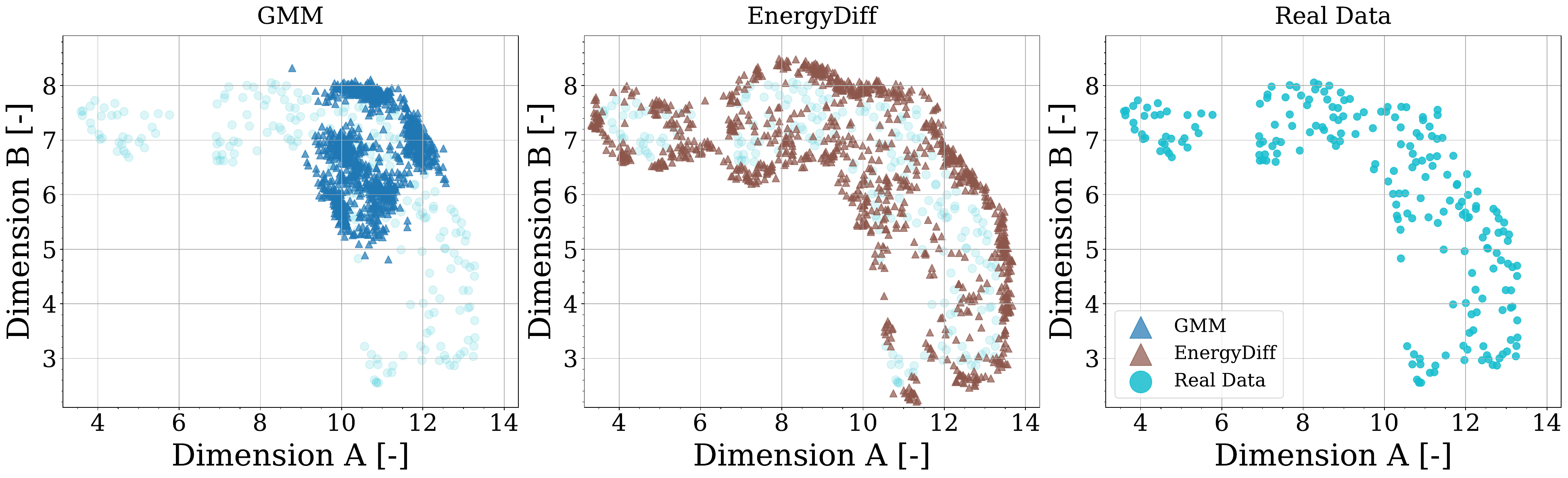}
    \caption{UMAP visualization of WPuQ transformer measurement data. Our model captures the complete manifold of the real data, while GMM fails to do so.}
    \label{fig:umap-wpuq-trafo}%
    \vspace{-2mm}
\end{figure}

\changed{While these metrics—GFD for comparing first and second moments, and KL, WD, and KS for evaluating marginal distributions—provide valuable insights, they do not reflect the complex temporal dependencies inherent in high-dimensional data distributions. Meanwhile, the temporal dependency is crucial in data generation tasks. For example, when a customer turns on an electrical appliance, it is likely the appliance will be kept on for a certain period of time instead of immediately shut down. Another example is if a customer takes a shower in the morning (peak heat consumption), it is unlikely they will do it again at night (no peak heat consumption).} Visualizing the temporal patterns of a 1440-step time series is difficult in general. Therefore, to better understand these high-dimensional time series, we use a dimensionality reduction tool, \changed{uniform manifold approximation and projection (UMAP)}~\cite{mcinnes2020UMAPUniformManifold}, to reduce the data into 2-dimensional. UMAP learns a manifold of the high-dimensional data and re-maps the manifold into the 2D space. As in Fig.~\ref{fig:umap-wpuq-trafo}, the data generated by \ours covers the whole manifold of the real data. Meanwhile, the figure demonstrates that our model does not merely replicate the training data points but instead interpolates and extrapolates data that align with the real manifold. However, data generated by GMM span only part of the manifold, indicating a discrepancy between GMM and the real data distribution. This suggests that using GMM synthetic data for energy system operation and planning could over-represent some scenarios and under-represent others, potentially leading to sub-optimal decisions. \changed{The small gap in the metric values and the large discrepancy in the manifolds between GMM and \ours proves that we cannot rely solely on the metric values to evaluate the quality of synthetic data.}

\begin{table}[t]
    \centering
    \caption{Evaluation Metric Results on WPuQ Transformer-level Residential Electricity Load Profiles}
    \scalebox{0.75}{
    \changed{\begin{tabular}{lcrrrrr}
        \toprule
        Model & Res. & MMD & GFD & KL & WD & KS \\
        \midrule
        GMM & \multirow{12}{*}{\SI{1}{\minute}} & $0.0043$ & $3.6560$ & $0.0120$ & $0.0129$ & $0.0275$ \\
         &  & $\pm~0.0005$ & $\pm~0.3434$ & $\pm~0.0028$ & $\pm~0.0043$ & $\pm~0.0101$ \\
        t-Copula &  & $0.0255$ & $23.1220$ & $0.0260$ & $0.0267$ & $0.0442$ \\
         &  & $\pm~0.0027$ & $\pm~1.9385$ & $\pm~0.0063$ & $\pm~0.0072$ & $\pm~0.0113$ \\
        VAE &  & $0.0410$ & $17.6446$ & $0.1680$ & $0.0663$ & $0.1330$ \\
         &  & $\pm~0.0041$ & $\pm~2.7142$ & $\pm~0.0196$ & $\pm~0.0065$ & $\pm~0.0149$ \\
        GAN &  & $0.6365$ & $177.8024$ & $1.2596$ & $0.2873$ & $0.6156$ \\
         &  & $\pm~0.0242$ & $\pm~4.9467$ & $\pm~0.0461$ & $\pm~0.0062$ & $\pm~0.0119$ \\
        \oursabbr &  & $0.0056$ & $3.4362$ & $0.0209$ & $0.0126$ & $0.0367$ \\
         &  & $\pm~0.0005$ & $\pm~0.1762$ & $\pm~0.0012$ & $\pm~0.0021$ & $\pm~0.0068$ \\
        \oursabbrcal &  & $\underline{0.0043}$ & $\underline{3.2535}$ & $\underline{0.0048}$ & $\underline{0.0111}$ & $\underline{0.0248}$ \\
         &  & $\pm~0.0007$ & $\pm~0.2249$ & $\pm~0.0018$ & $\pm~0.0049$ & $\pm~0.0098$ \\
        \midrule
        GMM & \multirow{12}{*}{\SI{15}{\minute}} & $0.0058$ & $0.1038$ & $0.0199$ & $0.0194$ & $0.0438$ \\
         &  & $\pm~0.0021$ & $\pm~0.0425$ & $\pm~0.0060$ & $\pm~0.0066$ & $\pm~0.0160$ \\
        t-Copula &  & $0.0055$ & $0.1652$ & $0.0139$ & $0.0149$ & $0.0314$ \\
         &  & $\pm~0.0008$ & $\pm~0.0384$ & $\pm~0.0034$ & $\pm~0.0068$ & $\pm~0.0173$ \\
        VAE &  & $0.0601$ & $1.5220$ & $0.2202$ & $0.0900$ & $0.1750$ \\
         &  & $\pm~0.0059$ & $\pm~0.2208$ & $\pm~0.0234$ & $\pm~0.0067$ & $\pm~0.0151$ \\
        GAN &  & $0.1976$ & $8.7108$ & $0.6575$ & $0.1189$ & $0.3274$ \\
         &  & $\pm~0.0121$ & $\pm~0.4893$ & $\pm~0.0261$ & $\pm~0.0084$ & $\pm~0.0046$ \\
        \oursabbr &  & $0.0084$ & $0.1597$ & $0.0220$ & $0.0217$ & $0.0456$ \\
         &  & $\pm~0.0017$ & $\pm~0.0567$ & $\pm~0.0046$ & $\pm~0.0096$ & $\pm~0.0122$ \\
        \oursabbrcal &  & $\underline{0.0040}$ & $\underline{0.0890}$ & $\underline{0.0108}$ & $\underline{0.0114}$ & $\underline{0.0244}$ \\
         &  & $\pm~0.0009$ & $\pm~0.0287$ & $\pm~0.0033$ & $\pm~0.0053$ & $\pm~0.0102$ \\
        \midrule
        GMM & \multirow{12}{*}{\SI{30}{\minute}} & $\underline{0.0037}$ & $\underline{0.0185}$ & $\underline{0.0137}$ & $\underline{0.0106}$ & $\underline{0.0261}$ \\
         &  & $\pm~0.0005$ & $\pm~0.0049$ & $\pm~0.0023$ & $\pm~0.0026$ & $\pm~0.0106$ \\
        t-Copula &  & $0.0051$ & $0.0658$ & $0.0183$ & $0.0160$ & $0.0287$ \\
         &  & $\pm~0.0010$ & $\pm~0.0169$ & $\pm~0.0031$ & $\pm~0.0059$ & $\pm~0.0066$ \\
        VAE &  & $0.0425$ & $0.6208$ & $0.1477$ & $0.0782$ & $0.1425$ \\
         &  & $\pm~0.0047$ & $\pm~0.1009$ & $\pm~0.0192$ & $\pm~0.0064$ & $\pm~0.0121$ \\
        GAN &  & $0.2333$ & $2.2553$ & $0.0702$ & $0.0438$ & $0.1105$ \\
         &  & $\pm~0.0093$ & $\pm~0.1104$ & $\pm~0.0069$ & $\pm~0.0062$ & $\pm~0.0121$ \\
        \oursabbr &  & $0.0082$ & $0.0630$ & $0.0262$ & $0.0195$ & $0.0469$ \\
         &  & $\pm~0.0015$ & $\pm~0.0406$ & $\pm~0.0044$ & $\pm~0.0083$ & $\pm~0.0158$ \\
        \oursabbrcal &  & $0.0046$ & $0.0324$ & $0.0175$ & $0.0145$ & $0.0344$ \\
         &  & $\pm~0.0015$ & $\pm~0.0152$ & $\pm~0.0042$ & $\pm~0.0071$ & $\pm~0.0174$ \\
        \midrule
        GMM & \multirow{12}{*}{\SI{1}{\hour}} & $0.0042$ & $0.0104$ & $0.0209$ & $0.0152$ & $0.0309$ \\
         &  & $\pm~0.0008$ & $\pm~0.0096$ & $\pm~0.0027$ & $\pm~0.0062$ & $\pm~0.0109$ \\
        t-Copula &  & $0.0043$ & $0.0229$ & $0.0267$ & $0.0123$ & $0.0341$ \\
         &  & $\pm~0.0005$ & $\pm~0.0035$ & $\pm~0.0034$ & $\pm~0.0037$ & $\pm~0.0142$ \\
        VAE &  & $0.3123$ & $1.6272$ & $0.3634$ & $0.1650$ & $0.3694$ \\
         &  & $\pm~0.0243$ & $\pm~0.0607$ & $\pm~0.0326$ & $\pm~0.0105$ & $\pm~0.0193$ \\
        GAN &  & $1.0355$ & $7.0019$ & $1.8982$ & $0.4753$ & $0.7511$ \\
         &  & $\pm~0.0366$ & $\pm~0.1938$ & $\pm~0.0795$ & $\pm~0.0090$ & $\pm~0.0094$ \\
        \oursabbr &  & $\underline{0.0039}$ & $0.0082$ & $\underline{0.0197}$ & $0.0123$ & $\underline{0.0297}$ \\
         &  & $\pm~0.0006$ & $\pm~0.0040$ & $\pm~0.0033$ & $\pm~0.0031$ & $\pm~0.0131$ \\
        \oursabbrcal &  & $0.0040$ & $\underline{0.0060}$ & $0.0249$ & $\underline{0.0119}$ & $0.0322$ \\
         &  & $\pm~0.0005$ & $\pm~0.0026$ & $\pm~0.0029$ & $\pm~0.0028$ & $\pm~0.0084$ \\
        \bottomrule
    \end{tabular}}}%
    \label{tab:wpuq-trafo-perf}%
    \vspace{-4mm}
\end{table}

\subsubsection{Residential PV Generation}
\begin{figure}[t]
    \centering
    \includegraphics[width=0.85\linewidth]{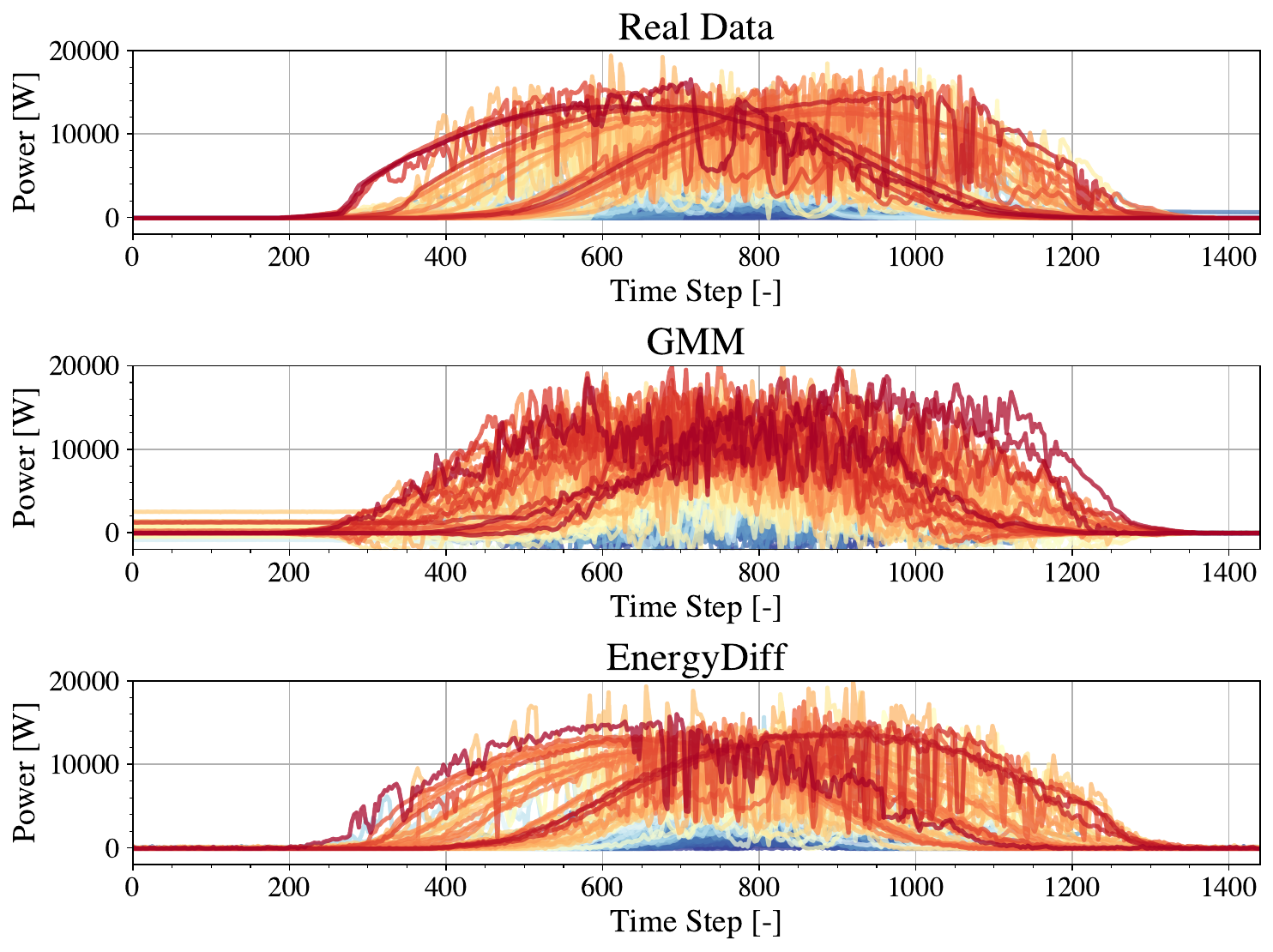}
    \caption{WPuQ PV generation curves in 1 minute resolution. GMM samples are overly noisy and contain unrealistic negative values.}
    \label{fig:samples-wpuq-pv-1min}%
    \vspace{-2mm}
\end{figure}

We show the numerical results for the WPuQ transformer\changed{-level} PV generation dataset in Table~\ref{tab:wpuq-pv-perf}. \changed{We observe that \ours still performs consistently well across different resolutions in most metrics by showing best or close to best scores. The exception is at the 1-minute resolution, where GMM has less than half of the MMD and GFD of \ours.} To further assess whether \changed{GMM indeed has better quality of generated data}, we visualize 100 real and synthetic samples from these two models in Fig.~\ref{fig:samples-wpuq-pv-1min}. It is noticed that both models capture the coarse wave shape of the data, i.e., \changed{(near)} zero generation during night time and peak generation around noon. \changed{However, according to the real data and physics common sense, the generation during nighttime should be absolute zero, while GMM sometimes violates this rule by having non-zero generation power at night.} \changed{Meanwhile,} GMM again shows unrealistic negative values. Most of the peak values of the real data lie around $15$kW, while the peak values of GMM data are overestimated. \changed{The data generated by GMM also appear to oscillate too much.}

\begin{figure}[t]
    \centering
    \includegraphics[width=0.92\linewidth]{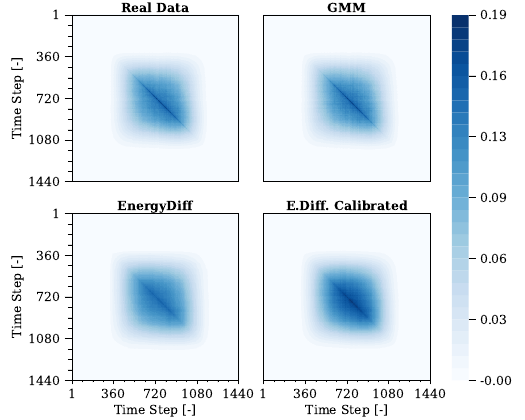}
    \caption{\changed{Covariance matrices of WPuQ PV generation data in 1 minute resolution. GMM is designed to match exactly the real data's covariance. \ours captures the same patterns in the covariance despite of small differences.}}
    \label{fig:cov-real-gmm-ddpm-ddpmc}
\end{figure}

\begin{figure}[t]
    \centering
    \includegraphics[width=\linewidth]{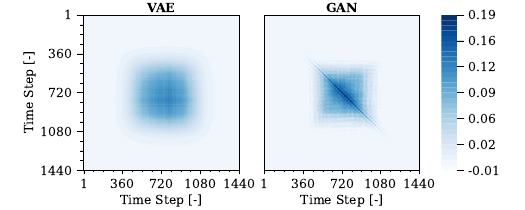}
    \caption{\changed{Covariance matrices of WPuQ PV generation data in 1 minute resolution from data generated by VAE and GAN. VAE captures the center but is not comparable to all the other models. GAN captures the covariance better but still falls short when compared to GMM and \ours.}}
    \label{fig:cov-vae-vaec-gan-ganc}
\end{figure}

\changed{Additionally, we visualize the covariance matrices of the real and generated data by each model in Fig.~\ref{fig:cov-real-gmm-ddpm-ddpmc} and Fig.~\ref{fig:cov-vae-vaec-gan-ganc}. As the figures illustrate, GMM indeed captures the covariance almost perfectly as it is designed to do so. \ours also captures the covariance well, both before and after the calibration. The calibration brings slight deterioration because it is not designed to optimize the covariance. However, the difference is small, which is guaranteed by \eqref{eq:ot}, as the calibration function is a minimizer of the transportation cost (the smallest changes possible on data) while achieving the calibration constraints.}

\begin{table}[t]
    \centering
    \caption{Evaluation Metric Results on WPuQ Transformer-level PV Generation}
    \scalebox{0.75}{
    \changed{\begin{tabular}{lcrrrrr}
        \toprule
        Model & Res. & MMD & GFD & KL & WD & KS \\
        \midrule
        GMM & \multirow{12}{*}{\SI{1}{\minute}} & $\underline{0.0009}$ & $\underline{1.3953}$ & $0.0336$ & $0.0136$ & $0.2637$ \\
         &  & $\pm~0.0002$ & $\pm~0.1069$ & $\pm~0.0099$ & $\pm~0.0012$ & $\pm~0.0018$ \\
        t-Copula &
                    & \multirow{2}{*}{\tripleast} & \multirow{2}{*}{\tripleast} & \multirow{2}{*}{\tripleast} & \multirow{2}{*}{\tripleast} & \multirow{2}{*}{\tripleast} \\
        &&&&&& \\
        VAE &  & $0.0242$ & $21.0066$ & $0.2380$ & $0.0326$ & $0.2493$ \\
         &  & $\pm~0.0018$ & $\pm~0.4760$ & $\pm~0.0305$ & $\pm~0.0014$ & $\pm~0.0022$ \\
        GAN &  & $0.0174$ & $19.8873$ & $0.0271$ & $0.0319$ & $0.4806$ \\
         &  & $\pm~0.0011$ & $\pm~0.5348$ & $\pm~0.0008$ & $\pm~0.0015$ & $\pm~0.0019$ \\
        \oursabbr &  & $0.0049$ & $5.5573$ & $0.1011$ & $0.0113$ & $0.2865$ \\
         &  & $\pm~0.0003$ & $\pm~0.1000$ & $\pm~0.0027$ & $\pm~0.0012$ & $\pm~0.0034$ \\
        \oursabbrcal &  & $0.0024$ & $4.8386$ & $\underline{0.0004}$ & $\underline{0.0031}$ & $\underline{0.0051}$ \\
         &  & $\pm~0.0004$ & $\pm~0.2178$ & $\pm~0.0002$ & $\pm~0.0024$ & $\pm~0.0028$ \\
        \midrule
        GMM & \multirow{12}{*}{\SI{15}{\minute}} & $0.0009$ & $\underline{0.0446}$ & $0.0413$ & $0.0134$ & $0.2607$ \\
         &  & $\pm~0.0002$ & $\pm~0.0092$ & $\pm~0.0349$ & $\pm~0.0013$ & $\pm~0.0035$ \\
        t-Copula &
                    & \multirow{2}{*}{\tripleast} & \multirow{2}{*}{\tripleast} & \multirow{2}{*}{\tripleast} & \multirow{2}{*}{\tripleast} & \multirow{2}{*}{\tripleast} \\
        &&&&&& \\
        VAE &  & $0.0114$ & $0.5784$ & $0.7152$ & $0.0257$ & $0.4526$ \\
         &  & $\pm~0.0009$ & $\pm~0.0257$ & $\pm~0.1454$ & $\pm~0.0013$ & $\pm~0.0021$ \\
        GAN &  & $0.0263$ & $1.4343$ & $0.0594$ & $0.0441$ & $0.4671$ \\
         &  & $\pm~0.0018$ & $\pm~0.0656$ & $\pm~0.0014$ & $\pm~0.0029$ & $\pm~0.0023$ \\
        \oursabbr &  & $0.0128$ & $0.3214$ & $0.4522$ & $0.0324$ & $0.4179$ \\
         &  & $\pm~0.0010$ & $\pm~0.0242$ & $\pm~0.0068$ & $\pm~0.0018$ & $\pm~0.0027$ \\
        \oursabbrcal &  & $\underline{0.0009}$ & $0.0652$ & $\underline{0.0011}$ & $\underline{0.0043}$ & $\underline{0.0057}$ \\
         &  & $\pm~0.0001$ & $\pm~0.0087$ & $\pm~0.0002$ & $\pm~0.0016$ & $\pm~0.0019$ \\
        \midrule
        GMM & \multirow{12}{*}{\SI{30}{\minute}} & $0.0009$ & $\underline{0.0136}$ & $0.0265$ & $0.0118$ & $0.2591$ \\
         &  & $\pm~0.0002$ & $\pm~0.0019$ & $\pm~0.0167$ & $\pm~0.0006$ & $\pm~0.0027$ \\
        t-Copula &
                    & \multirow{2}{*}{\tripleast} & \multirow{2}{*}{\tripleast} & \multirow{2}{*}{\tripleast} & \multirow{2}{*}{\tripleast} & \multirow{2}{*}{\tripleast} \\
        &&&&&& \\
        VAE &  & $0.0062$ & $0.2638$ & $0.2235$ & $0.0180$ & $0.3253$ \\
         &  & $\pm~0.0004$ & $\pm~0.0119$ & $\pm~0.0093$ & $\pm~0.0010$ & $\pm~0.0014$ \\
        GAN &  & $0.0140$ & $0.3264$ & $0.0649$ & $0.0373$ & $0.2473$ \\
         &  & $\pm~0.0025$ & $\pm~0.0405$ & $\pm~0.0015$ & $\pm~0.0047$ & $\pm~0.0024$ \\
        \oursabbr &  & $0.0101$ & $0.1315$ & $0.3658$ & $0.0264$ & $0.3749$ \\
         &  & $\pm~0.0016$ & $\pm~0.0188$ & $\pm~0.0039$ & $\pm~0.0032$ & $\pm~0.0027$ \\
        \oursabbrcal &  & $\underline{0.0009}$ & $0.0211$ & $\underline{0.0020}$ & $\underline{0.0033}$ & $\underline{0.0050}$ \\
         &  & $\pm~0.0001$ & $\pm~0.0029$ & $\pm~0.0001$ & $\pm~0.0015$ & $\pm~0.0015$ \\
        \midrule
        GMM & \multirow{12}{*}{\SI{1}{\hour}} & $0.0009$ & $\underline{0.0060}$ & $0.0186$ & $0.0110$ & $0.2449$ \\
         &  & $\pm~0.0001$ & $\pm~0.0027$ & $\pm~0.0122$ & $\pm~0.0006$ & $\pm~0.0018$ \\
        t-Copula &
                    & \multirow{2}{*}{\tripleast} & \multirow{2}{*}{\tripleast} & \multirow{2}{*}{\tripleast} & \multirow{2}{*}{\tripleast} & \multirow{2}{*}{\tripleast} \\
        &&&&&& \\
        VAE &  & $0.0041$ & $0.0876$ & $0.2703$ & $0.0161$ & $0.3125$ \\
         &  & $\pm~0.0003$ & $\pm~0.0040$ & $\pm~0.1000$ & $\pm~0.0017$ & $\pm~0.0019$ \\
        GAN &  & $0.0090$ & $0.1173$ & $0.1129$ & $0.0369$ & $0.4429$ \\
         &  & $\pm~0.0015$ & $\pm~0.0122$ & $\pm~0.0030$ & $\pm~0.0034$ & $\pm~0.0018$ \\
        \oursabbr &  & $\underline{0.0008}$ & $0.0086$ & $0.0687$ & $0.0045$ & $0.2081$ \\
         &  & $\pm~0.0001$ & $\pm~0.0041$ & $\pm~0.0017$ & $\pm~0.0009$ & $\pm~0.0026$ \\
        \oursabbrcal &  & $0.0008$ & $0.0094$ & $\underline{0.0038}$ & $\underline{0.0037}$ & $\underline{0.0083}$ \\
         &  & $\pm~0.0002$ & $\pm~0.0045$ & $\pm~0.0004$ & $\pm~0.0019$ & $\pm~0.0021$ \\
        \bottomrule
    \end{tabular}}}%
    \label{tab:wpuq-pv-perf}%
    \vspace{-2mm}
\end{table}

\subsection{Marginal Calibration Evaluation}
\label{sec:marginal-calibration}
\changed{As Tables~\ref{tab:wpuq-perf} to \ref{tab:wpuq-pv-perf} illustrate, the proposed Marginal Calibration consistently improves the overall results of all metrics. The marginal-based metrics, KL, WD, and KS, are almost always improved and reduced to near-zero values, which is exactly the design goal. In the cases where these metrics are not improved, the pre-calibration data are already very good, and the post-calibration data would have similar metric scores.}

\begin{figure*}[t]
    \centering
    \includegraphics[width=0.85\linewidth]{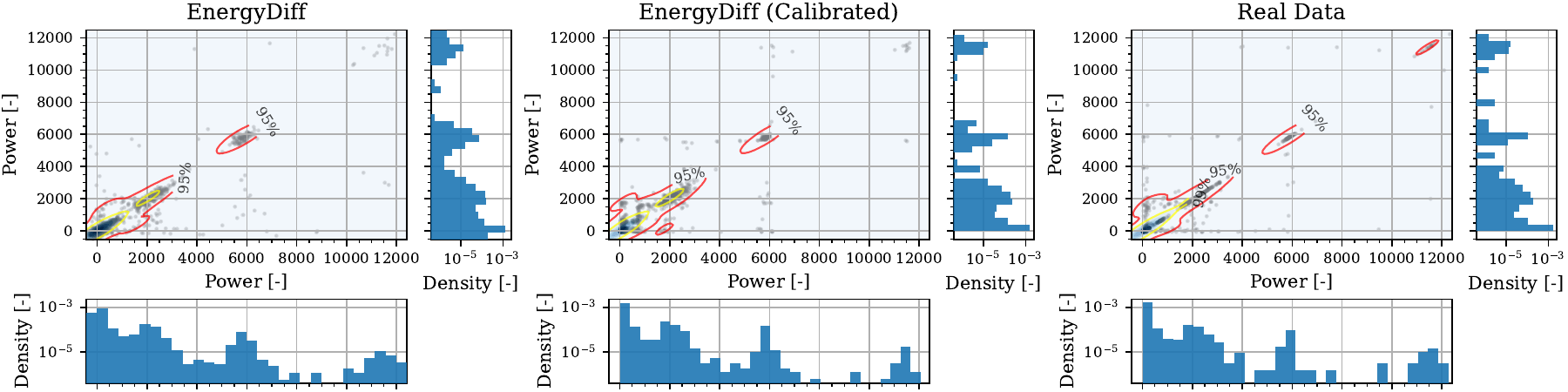}
    \caption{\textbf{Top left} is a scatter plot of the real and generated heat pump data by \ours~at the $115$-th and $116$-th step, before and after the marginal calibration. Gray dots are data points. Contour lines are based on based on Gaussian kernel density estimation. \textbf{Bottom} and \textbf{Top right} are the corresponding marginal distributions. The proposed calibration applies the minimum change to the data but makes the marginal distributions almost exactly match the real data.}
    \label{fig:calibrate-before-after}%
    \vspace{-2mm}
\end{figure*}

Next, we present \changed{more details of} the effect of the proposed Marginal Calibration \changed{by} randomly \changed{selecting} two consecutive time steps to visualize the temporal pattern differences in Fig.~\ref{fig:calibrate-before-after}. Here, we have selected the $115$-th and $116$-th time steps of the WPuQ heat pump data, corresponding to \texttt{01:55} and \texttt{01:56} midnight. We observe small changes in the scatter plot in the middle. However, the marginal distributions are brought closer to the real data. This is because our calibration is based on OT, \changed{as in \eqref{eq:ot}}, which minimizes the changes but guarantees the calibrated data has the exact CDF of the training data. The remaining mismatch of the marginal distributions is due to the ECDF estimation error from the training data. However, \changed{from the previous numerical results, we see that} this mismatch is small \changed{in practice, including the case of limited $408$ training sample size in Table~\ref{tab:cossmic-pv-residential-perf}}.

\subsection{\changed{Computation Time and Stability}}
\label{sec:computation-time}

\begin{table}[t]
    \centering
    \caption{\changed{Computation Time/Stability Comparison of Generating 4000 Samples.}}
    \scalebox{0.75}{%
    \changed{\begin{tabular}{ccccccc}
    \toprule
    \multirow{2}{*}{Time/Conv.} & Training & Sampling & Convergence & Training & Sampling & Convergence \\
    & Time & Time & Percentage & Time & Time & Percentage \\
    \midrule
    Resolution & \multicolumn{3}{c}{1-minute} & \multicolumn{3}{c}{1-hour} \\
    \midrule
    GMM & \SI{1}{\minute} & $<$\SI{10}{\second} & \SI{100}{\percent} & \SI{1}{\minute} & $<$\SI{10}{\second} & \SI{100}{\percent} \\
    t-Copula & \SI{15}{\minute} & \SI{5}{\minute} & \SI{20}{\percent} & \SI{5}{\minute} & \SI{1}{\minute} & \SI{66.67}{\percent} \\
    VAE & \SI{30}{\minute} & $<$\SI{10}{\second} & \SI{100}{\percent} & \SI{20}{\minute} & $<$\SI{10}{\second} & \SI{100}{\percent} \\
    GAN & \SI{3}{\hour} & $<$\SI{10}{\second} & \SI{60}{\percent} & \SI{1}{\hour} & $<$\SI{10}{\second} & \SI{83.33}{\percent} \\
    \oursabbr$^{(1000)}$ & \SI{3}{\hour} & \SI{70}{\minute} & \SI{100}{\percent} & \SI{1}{\hour} & \SI{5}{\minute} & \SI{100}{\percent} \\
    \oursabbr$^{(100)}$ & \SI{3}{\hour} & \SI{8}{\minute} & \SI{100}{\percent} & \SI{1}{\hour} & \SI{30}{\second} & \SI{100}{\percent} \\
    \bottomrule
    \multicolumn{7}{l}{\oursabbr$^{(1000)}$ \ours with 1000 steps ancestral sampling.}\\
    \multicolumn{7}{l}{\oursabbr$^{(100)}$ \ours with 100 steps accelerated DPM-Solver~\cite{lu2022DPMSolverFastODE} sampling.}
    \end{tabular}}%
    }
    \label{tab:computation-time-comparison}
\end{table}

\begin{table}
    \centering
    \caption{\changed{Computation Time and Memory Consumption w/o Folding}}
    \scalebox{0.75}{%
    \changed{\begin{tabular}{ccccc}
    \toprule
     & Forward Time & Backward Time & Total Time & Peak GPU Memory  \\
     &  (ms) &  (ms) & (ms) & (GiB) \\
    \midrule
    No Folding & $511.41\pm 28.28$ & $350.44\pm 3.81$ & $861.85\pm 30.20$ & 28.76 \\
    $r=8$ & $410.90\pm 0.87$ & $157.17\pm 5.92$ & $565.00\pm 6.34$ & 2.03 \\
    \bottomrule
    \end{tabular}}}
    \label{tab:folding-time-mem}
\end{table}

The training time and difficulty of GMM, t-Copula, VAE, GAN, and \ours varies significantly. All \ours, VAE, and GAN experiments, including training and generation, are done on a single-card system with NVIDIA A100. GMM and t-Copula experiments are carried out on a system with two Intel Xeon Platinum 8360Y CPUs with in total 72 cores. We summarize the computation time of tested models on 1-minute and 1-hour data resolution in Table~\ref{tab:computation-time-comparison}. \ours is trained with $S=4000$ diffusion steps (no impact on training time). In practice, we always use DPM-Solver~\cite{lu2022DPMSolverFastODE} to accelerate the sampling at 100 sampling steps. The performance degradation compared with $4000$ sampling steps (no acceleration) and $1000$ sampling step (simplest acceleration~\cite{nichol2021ImprovedDenoisingDiffusion}) is negligible, so we do not include those results. We see the GANs models that have significantly larger GFD as not converged. \changedii{For GMM and t-Copula models, we conclude their divergence if they do not converge after three hours of retrying with different initialization.}

\ours training requires approximately 3 hours for 1-minute resolution data and 1 hour for 1-hour resolution data. Generating $4000$ instances of the 1-minute data from \ours takes around 8 minutes, while it only takes seconds for 1-hour data. The GMM is more computation-efficient, with both fitting and sampling taking less than 1 minute. Fitting the t-Copula model for the 1-minute resolution takes approximately 15 minutes but frequently fails due to the numerical instability of the optimization process. Notably, in several datasets like the CoSSMic residential electricity load profile, t-Copula failed to converge despite multiple tries. We observe that this instability is particularly exacerbated in high-resolution data, such as 1 minute, and when the training set size is thousands or larger. Overall, \ours, VAE, and GMM are computationally robust. \changed{We acknowledge GMM's computation efficiency advantage over \ours. The extended sampling time, inherent to diffusion models, results from their long Markov chain during the reverse process. However, synthetic data applications rarely demand real-time generation. Data quality is often prioritized over computational efficiency. For example, in downstream machine learning tasks such as reinforcement learning, synthetic data generation occurs during the offline training phase, where longer data generation processes are acceptable, while compromised data quality directly impacts model performance. Furthermore, the acceleration methodologies of diffusion models are going through rapid advancements. DPM-Solver~\cite{lu2022DPMSolverFastODE} demonstrates sampling speedups of up to 20 times with little performance degradation, as ongoing research promises further improvements.}

\changed{In addition, we show the effect of the folding operation in Table~\ref{tab:folding-time-mem}. We present the computation time of a single forward pass and a single backward pass, and the peak GPU memory usage when applying \ours on 1-minute daily heat pump consumption data, i.e., 1440-step time series. Not only more than \SI{34}{\percent} of computation time is saved, but more importantly, the GPU memory consumption is significantly reduced with folding. In Table~\ref{tab:folding-time-mem}, the batch size is set to $16$. On our platform with NVIDIA A100 with \SI{40}{\giga \byte} graphics memory, the maximum batch size is around $22$ without folding because the memory use is nearly proportional to the batch size. Meanwhile, folding with $r=8$ allows us to use a batch size as large as $315$, which improves the training efficiency significantly. The extremely large memory consumption without folding also limits the model scale. Consequently, the folding operation is necessary for high-resolution data in terms of practical computation time and memory concerns.}

\section{Conclusion}
DDPMs are powerful generative models that have become the most popular choice in the image and audio generation domain. However, the standard DDPM has high computation and memory complexity related to the input data size, making them unsuitable for generating high-resolution time series data such as 1440-step 1-minute daily load profiles. Additionally, despite their capability to capture complex dependencies, they do not necessarily yield precise marginal distributions, which is crucial for accurately representing high-consumption or high-generation scenarios. To address these issues, we proposed \ours, a DDPM-based universal energy time series generation framework. With a tailored denoising process, \ours generates high-quality data across different energy domains at various time resolutions and both the customer and transformer levels. Our proposed Marginal Calibration technique ensures that \ours captures precise marginal distributions. 

\changed{Building upon these results, three promising directions for future work emerge. First, a more sophisticated training procedure can be developed by incorporating multiple data sources and resolutions within the same energy domain. Second, the \ours framework can be extended to support conditional generation. By conditioning on labels such as annual consumption, weather patterns, and seasonal indicators, the model could extrapolate data to previously unobserved combinations of these conditions. Such conditioning models can also enable practical applications like inpainting missing data segments and super-resolution from low-resolution measurements. At last, since our model is a general framework for time-series, it is worth investigating its performance on time series with a longer span, such as cross-day data and monthly and yearly data.}

\bibliographystyle{IEEEtran}
\bibliography{energy_system_planning, energy_transition, machine_learning}

\begin{IEEEbiography}[{\includegraphics[width=1in,height=1.25in,clip,keepaspectratio]{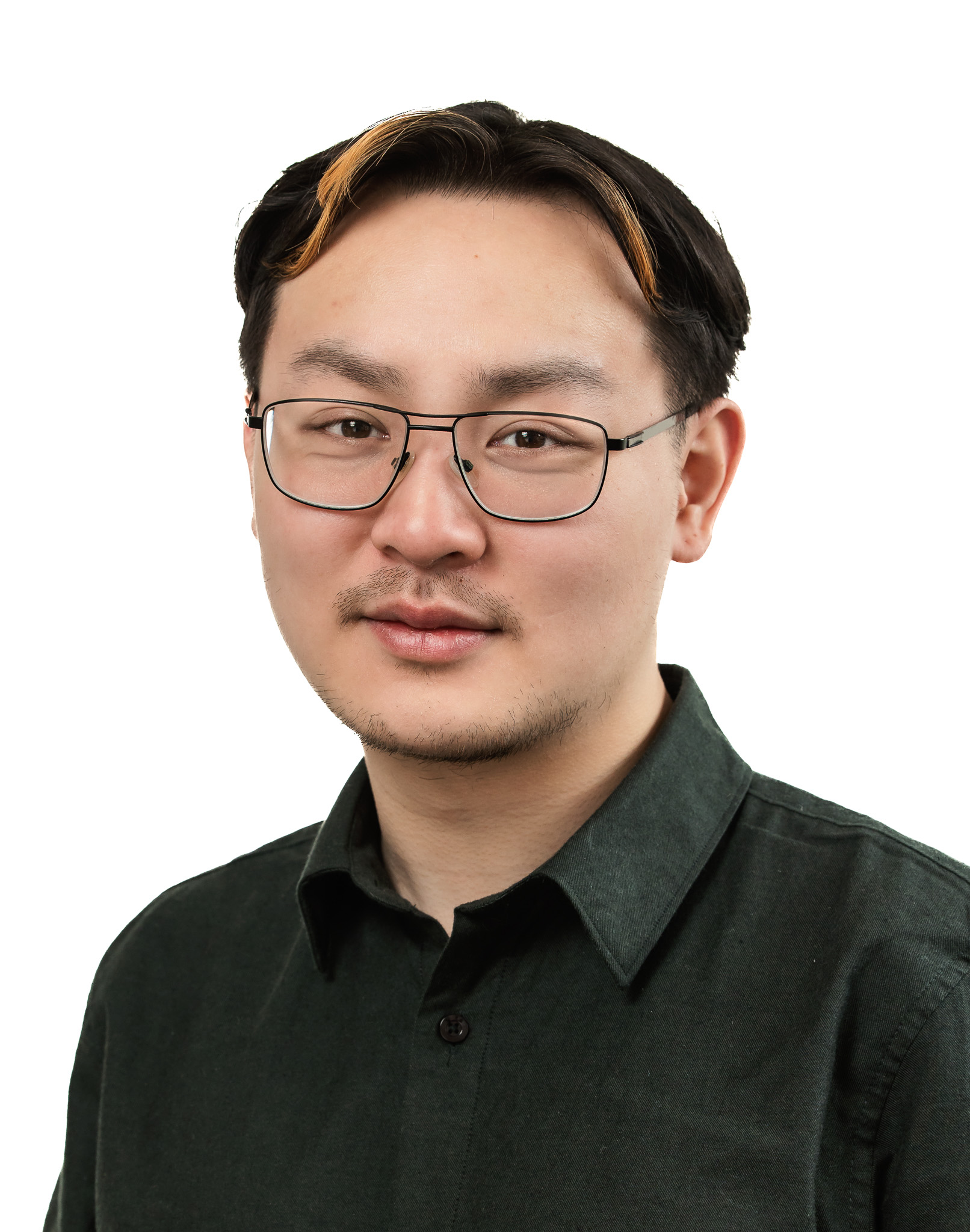}}]{Nan Lin}
is a Ph.D. candidate in the Intelligent Electrical Power Grids (IEPG) Group at Delft University of Technology (TU Delft), the Netherlands. He received his bachelor degree in Electrical Engineering in 2020 at Xi'an Jiaotong University, Shaanxi, China. In Aug. 2022, he received his M.Sc. degree in Electrical Engineering with the Signal Processing Systems (formerly known as Circuits and Systems) Group at TU Delft. His research interests are primarily in machine learning or deep learning from a probabilistic perspective. More specifically, he is interested in Bayesian machine learning, copula models, deep generative models, machine learning on graphs, transfer learning, and reinforcement learning. Currently, he is focusing on the applications of state-of-the-art machine learning techniques in power systems.%
\end{IEEEbiography}

\begin{IEEEbiography}%
[{\includegraphics[width=1in,height=1.25in,clip,keepaspectratio]%
{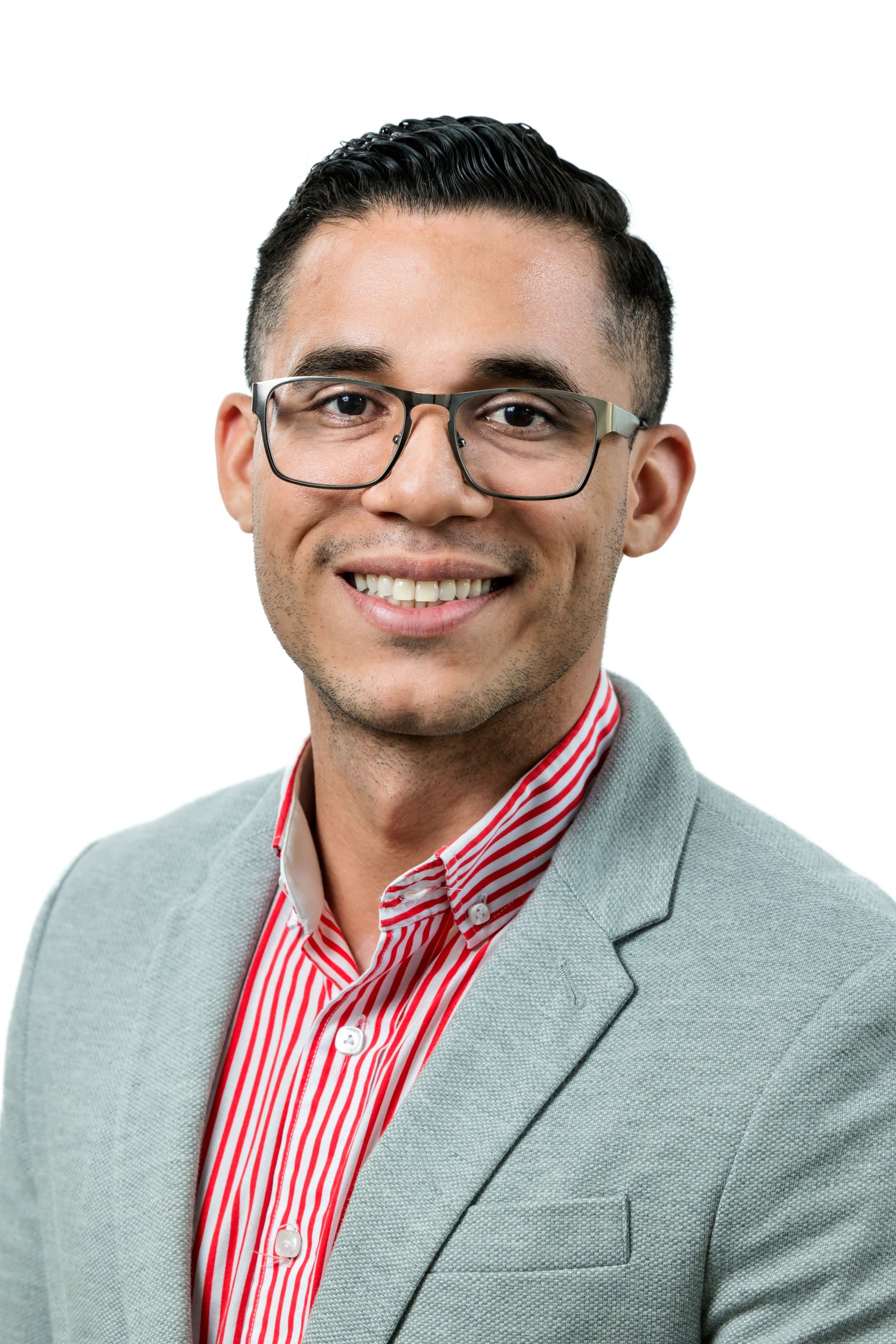}}]{Pedro P. Vergara}
(M'19) was born in Barranquilla, Colombia in 1990. He received the B.Sc. degree (with honors) in electronic engineering from the Universidad Industrial de Santander, Bucaramanga, Colombia, in 2012, and the M.Sc. degree in electrical engineering from the University of Campinas, UNICAMP, Campinas, Brazil, in 2015. In 2019, he received his Ph.D. degree from the University of Campinas, UNICAMP, Brazil, and the University of Southern Denmark, SDU, Denmark, funded by the Sao Paulo Research Foundation (FAPESP).  In 2019, he joined the Eindhoven University of Technology, TU/e, in The Netherlands as a Postdoctoral Researcher. In 2020, he was appointed Assistant Professor at the Intelligent Electrical Power Grids (IEPG) group at Delft University of Technology, also in The Netherlands. His main research interests include the development of algorithms for the control, planning, and operation of electrical distribution systems with high penetration of low-carbon energy resources (e.g, electrical vehicles, PV systems, electric heat pumps) using optimization and machine learning approaches. Dr. Vergara received the Best Presentation Award at the Summer Optimization School in 2018 organized by the Technical University of Denmark (DTU) and the Best Paper Award at the 3rd IEEE International Conference on Smart Energy Systems and Technologies (SEST), in Turkey, in 2020.
\end{IEEEbiography}

\begin{IEEEbiography}[{\includegraphics[width=1in,height=1.25in,clip,keepaspectratio]{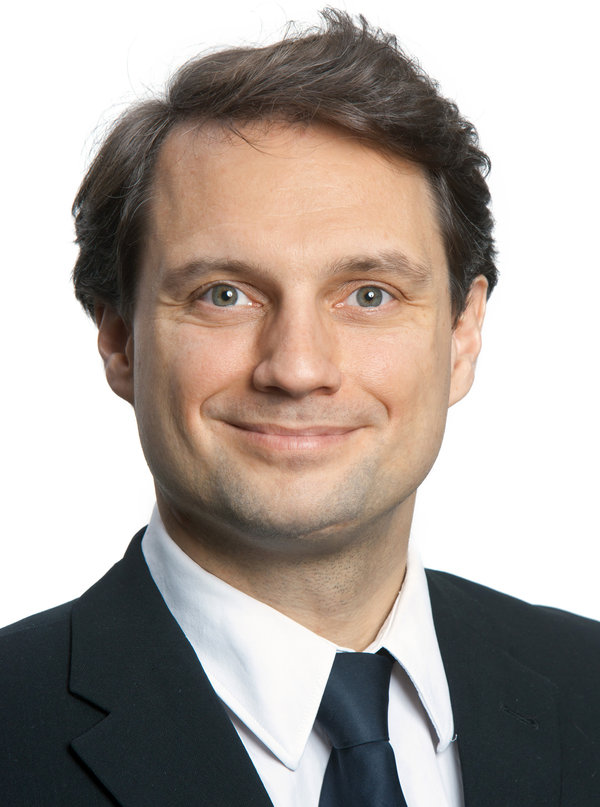}}]{Peter Palensky}
received the  M.Sc. degree in electrical engineering and the Ph.D.  and Habilitation degrees from the Vienna University
of Technology, Austria, in 1997, 2001, and 2015, respectively. He co-founded Envidatec, a German startup on energy management and analytics. In 2008, he joined the Lawrence Berkeley National Laboratory, Berkeley, CA, USA, as a Researcher, and the University of Pretoria, South Africa. In 2009, he became appointed as the Head of the Business Unit, Austrian Institute of Technology (AIT) in sustainable building technologies, where he was the first Principal Scientist of Complex Energy Systems. In 2014, he was appointed as a Full Professor in intelligent electric power grids with TU Delft, The Netherlands. He is active in international committees, such as ISO or CEN. His research interests include energy automation networks, smart
grids, and modeling intelligent energy systems. He also serves as an IEEE IES AdCom Member-at-Large in various functions for IEEE. He is the past Editor-in-Chief of IEEE Industrial Electronics Magazine and an associate editor of
several other IEEE publications and regularly organizes IEEE conferences.
\end{IEEEbiography}

\end{document}